\documentclass[11pt]{article}

\usepackage[preprint]{acl}

\usepackage{times}
\usepackage{latexsym}
\usepackage{url}

\usepackage[T1]{fontenc}

\usepackage[utf8]{inputenc}

\usepackage{microtype}

\usepackage{inconsolata}

\usepackage{graphicx}

\usepackage{comment}
\usepackage[inline]{enumitem}
\usepackage{soul}

\usepackage{amsfonts}
\usepackage{booktabs}
\usepackage{makecell}
\usepackage{subcaption}
\usepackage[most]{tcolorbox}
\AtBeginEnvironment{tabular}{\footnotesize} 
\AtBeginEnvironment{tcolorbox}{\small}

\usepackage{multirow}

\usepackage{siunitx} 

\title{Beyond Marginal Distributions: A Framework to Evaluate the Representativeness of Demographic-Aligned LLMs}

\author{
 \textbf{Tristan Williams\textsuperscript{1}}\quad \quad 
 \textbf{Franziska Weeber\textsuperscript{2}}\quad \quad 
 \textbf{Sebastian Padó\textsuperscript{2}}\quad\quad 
 \textbf{Alan Akbik\textsuperscript{1}}
\\
\\
 \textsuperscript{1}Humboldt University of Berlin \quad  \quad
  \textsuperscript{2}University of Stuttgart
\\
 \small{
   \texttt{tdw75@outlook.com, \{franziska.weeber|pado\}@ims.uni-stuttgart.de, alan.akbik@hu-berlin.de}
}
}

\begin{document}
\setlength{\abovedisplayskip}{5pt}
\setlength{\abovedisplayshortskip}{2pt}
\setlength{\belowdisplayskip}{8pt}
\maketitle
\begin{abstract}

Large language models are increasingly used to represent human opinions, values, or beliefs, and their steerability towards these ideals is an active area of research. Existing work focuses predominantly on aligning marginal response distributions, treating each alignment evaluation example independently. While essential, this may overlook deeper latent structures that characterise real populations and underpin cultural values theories. We propose a framework for evaluating the \textit{representativeness} of aligned models through multivariate correlation patterns in addition to marginal distributions. We show the value of our evaluation scheme by comparing two model steering techniques (persona prompting and demographic fine-tuning) and evaluating them against human responses from the World Values Survey. While the demographic fine-tuned model better approximates marginal response distributions, persona prompting performs marginally better at reproducing the empirical correlation structure between survey items. Despite this reversal, neither technique aligns with human correlation patterns. We conclude that representativeness is a distinct aspect of value alignment and an evaluation focused on marginals can mask structural failures, leading to overly optimistic conclusions about model representativeness. 
\end{abstract}

\section{Introduction}

As Large Language Models (LLMs) are increasingly integrated into socially sensitive domains, the challenge of pluralistic alignment, namely ensuring AI reflects the diverse intentions, ethical norms, and beliefs of a heterogeneous global population, has moved to the forefront of research \citep{gabriel_artificial_2020, weidinger_taxonomy_2022, sorensen_position_2024, lu_model_2024}. A central difficulty for alignment lies in the non-monolithic nature of human values \citep{gabriel_artificial_2020}. Attempts to align AI systems with a monolithic conception of human values risk marginalising minority perspectives rather than preserving their inherent heterogeneity.

\begin{figure}[tb!]
    \centering
    \includegraphics[width=1.0\linewidth]{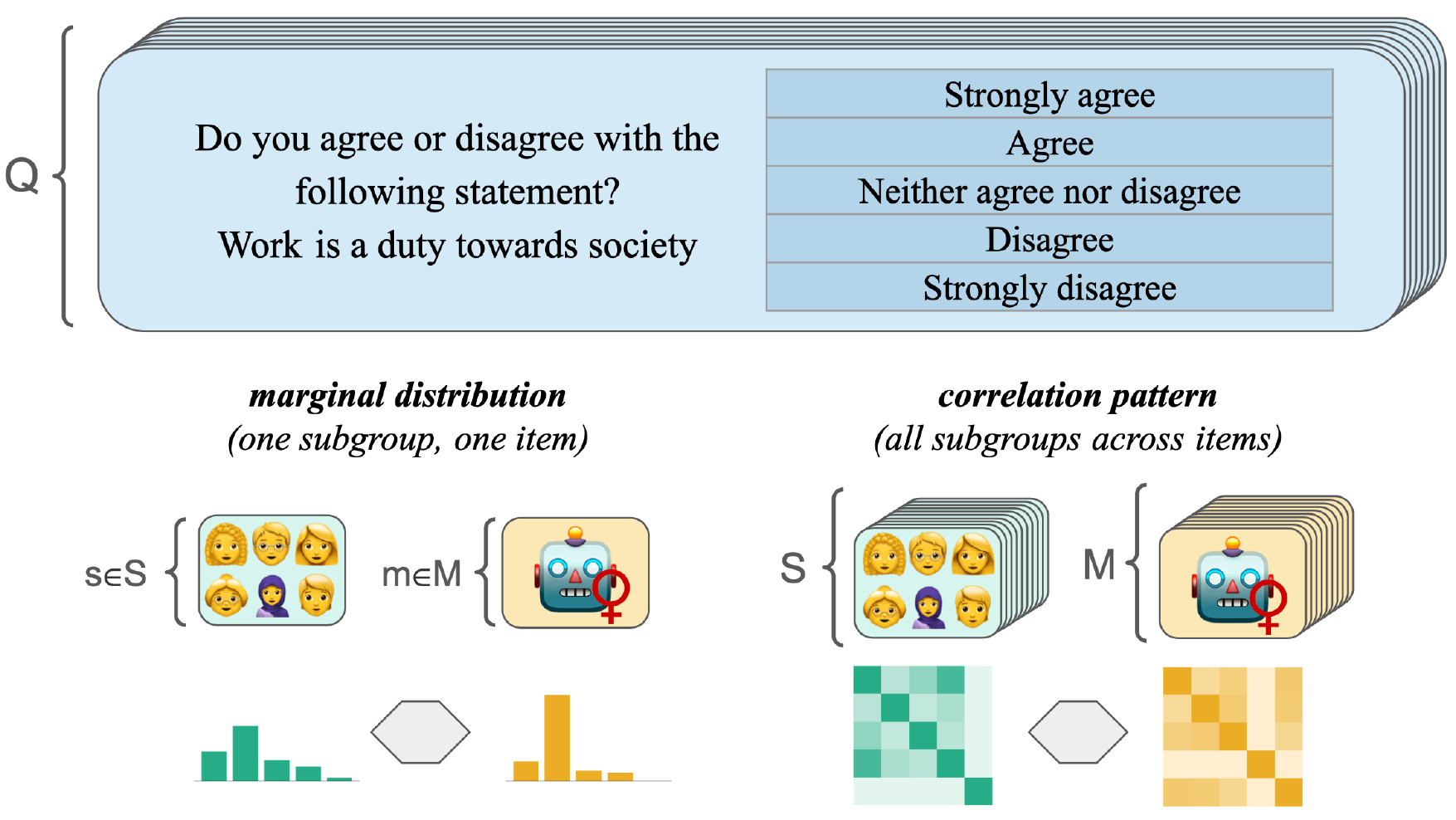}
    \caption{Overview of our suggested framework. Given a set of survey questions $Q$ (top), we compare marginal responses of one human subgroup $s \in S$ and a steered model $m \in M$ for each question (left) as well as the correlation structures across subgroups $S$ and across models $M$ over all questions (right).}
    \label{fig:overview}
\end{figure}

This paper engages directly with this problem by presenting a new methodology for evaluating \textit{representativeness} using the popular medium of value surveys \citep{argyle_out_2023, santurkar_whose_2023, bisbee_synthetic_2024, durmus_towards_2024, suh_language_2025}. We argue that the current focus in the literature on marginal distributions, i.e., comparing the response distributions of isolated questions, is a low bar for success. A truly representative model must capture not only what people think on each question individually, but also how their opinions are structured across questions. As noted in other fields, such as psycholinguistics \citep{Hollis2017}, focusing on isolated averages can mask significant structural failures. For example, a model might correctly approximate the support for two different policies independently, but fail to capture the fact that, in real populations, support for one is highly correlated with opposition to the other. Figure~\ref{fig:overview} represents our evaluation framework. \footnote{Code available at \url{https://github.com/tdw75/beyond-marginal-distributions}}

Using the World Values Survey (WVS) as our ground truth, we apply this correlation-based framework to compare two methods for model steering (i.e., conditioning models to become more representative of various demographic subgroups): \begin{enumerate*}[label=(\arabic*)]
    \item persona prompting with general-purpose LLMs; and
    \item demographically fine-tuned models, specifically OpinionGPT \citep{haller_opiniongpt_2024}
\end{enumerate*}. Persona prompting uses prompt engineering to alter the model's input context and invoke its internal "concept" of a demographic from its pretrained distribution. In contrast, fine-tuning embeds a group's perspectives directly into the model's parameters by training on natural language corpora authored by members of that group. 

Our primary contribution is the introduction of a new framework for survey-based evaluations of LLMs that uses both marginal distributions and correlation structures to reveal hidden failures in model representativeness and is easy to adapt to other use cases as required. As a secondary contribution, we demonstrate the diagnostic value of our framework through an illustrative case study comparing demographic fine-tuned and persona-prompted LLMs for model steering.

Using this framework, we investigate the following research questions: \begin{enumerate}[label=RQ\arabic*, leftmargin=*] 
    \item To what extent can common steering methods shift marginal response distributions toward those observed in human survey data? \label{rq1} 
    \item Do improvements in marginal alignment imply improved structural alignment in terms of inter-question correlation patterns? \label{rq2} 
    \item How does the representativeness vary across demographic subgroups and value domains when evaluated under this framework? \label{rq3} 
\end{enumerate}

We find that while fine-tuning and persona prompting can move models towards "human-like" marginals, substantial divergence in the correlation structure remains under both approaches.

\section{Related Work}

\subsection{Representativeness in LLMs}
Empirical studies have found LLMs to be more representative of some demographic groups than others. 
Geographically, they tend to align more closely with Western values \citep{durmus_towards_2024, tao_cultural_2024}. \citet{santurkar_whose_2023} also find representativeness disparities by education, religion, and socioeconomic status, where the opinions of Open\-AI's instruction-tuned models align better with those of the annotating group's demographics.
 
Politically, LLMs have been shown to represent left-leaning opinions
\citep{perez_discovering_2022, hartmann_political_2023, martin_ethico-political_2023, feng_pretraining_2023, fujimoto_revisiting_2023, weeber_political_2025}. 
\citet{ceron_beyond_2024} found that while most models do display \textit{political bias} (the favouring of specific policy positions), there is insufficient evidence of a \textit{political worldview} (a consistent ideological orientation across domains).

Even studies claiming representative results in LLMs should be treated with caution, with many conclusions based on demographically narrow evaluations \citep{sen_missing_2025}. 
In addition to the general misalignment, the distribution of true human responses shows greater diversity than that of model responses 
\citep{durmus_towards_2024, santurkar_whose_2023}. This is particularly a problem  in instruction-tuned models, in line with theoretical limitations of RLHF \citep{kirk_understanding_2024, xiao_algorithmic_2024}.

\subsection{Measuring Representativeness}

Most studies on human-LLM alignment generate responses independently at the survey item level and assess representativeness by comparing the marginal response distributions through statistical metrics (response means or variances) and distance/divergence metrics of answer distributions (Jenson-Shannon, Wasserstein, Kullblack-Leibler, etc.) \citep{bisbee_synthetic_2024, argyle_out_2023, santurkar_whose_2023, durmus_towards_2024, tao_cultural_2024, sen_missing_2025, gupta_bias_2024}. While these metrics are essential, they treat each question individually. 

In contrast, work in the social sciences considers underlying multivariate value patterns to be central. Examples are classic cultural value frameworks such as the Inglehart-Weizel Cultural Map \citep{inglehart_globalization_2000, inglehart_inglehart-welzel_2023}, Hofstede's Cultural Dimensions \citep{hofstede_cultures_1980}, and Schwartz's Theory of Basic Human Values \citep{schwartz_universals_1992}. True representativeness therefore also requires the preservation of the underlying structures that constitute cultural dimensions. Recent work by \citet{munker_fingerprinting_2025} proposes a similar framework to ours, also based on correlation structure and  termed \textit{fingerprinting}, for analysing LLM-generated survey responses demonstrated on a single psychometric instrument. In contrast, our work sees correlation structures as a necessary criterion for population-level representativeness and value alignment with human survey data.

\subsection{Steering and Aligning LLMs}
Model steering aims to counteract the collapse toward an "average human preference" observed in many modern LLMs \citep{sorensen_position_2024}. Instead, they seek to explicitly push model outputs to reflect the viewpoints of a specific demographic, a process aligned with the goal of distributional pluralistic alignment. In the context of this paper, we define \textbf{\textit{model steering}} as any method that attempts to encourage a language model's outputs to more closely reflect a specific target distribution. This can be achieved in different ways. We discuss two: 

\textbf{\textit{Persona prompting}} is widely used to steer a model towards the perspective of a demographic subgroup using prompts that define attributes or an identity, i.e., a persona. As a form of prompt engineering as it does not require retraining and can therefore be applied to almost any available LLM without having access to model weights or output probabilities. While there has been evidence of limited steerability toward the intended target groups \citep{santurkar_whose_2023, argyle_out_2023, bisbee_synthetic_2024, durmus_towards_2024}, in many cases the resulting distributions could still not be considered to be representative. Outputs fail to capture the intra-group heterogeneity found in true responses \citep{santurkar_whose_2023, bisbee_synthetic_2024} and often rely on reductive, stereotypical or even harmful associations \citep{durmus_towards_2024, gupta_bias_2024}. Therefore, it is unclear whether persona prompting as a steering technique can truly capture the diversity of human subgroups rather than merely activating surface-level semantic features.

\textbf{\textit{Fine-tuning}}
is an alternative to prompt-based model steering by adjusting the model weights or training an adapter \cite{hu_lora_2021} on data that is representative of the group the model should be aligned to. \textit{OpinionGPT} \citep{haller_opiniongpt_2024} uses Reddit data to create adapters for different genders, geographic regions, age groups, and political views. To align models with more left- or right-leaning political views, \citet{feng_pretraining_2023} use data from Reddit and news outlets while \citet{weeber_political_2025} use political manifestos. \textit{CultureLLM} \citep{li_culturellm_2025} and \textit{SubPOP} \citep{suh_language_2025} leverage fine-tuning on public opinion surveys for task-specific improvements and improved cultural alignment. Although less flexible than prompting, results from fine-tuning suggest that internalising subgroup perspectives through parameter adaptation may yield more stable, generalisable, and representative models. While fine-tuning requires computational effort and access to model weights, existing models can be reused, as we do in this work.

\section{Evaluation Framework}\label{sec:framework}

To assess how representative generated responses are of true human responses, our approach considers both the marginal response distributions as well as the correlation structures. Each comparison addresses a different question:
\begin{enumerate}[leftmargin=*, noitemsep]
    \item \textbf{\textit{Marginal distributions:}} How well do simulated responses approximate the ground truth response distribution? How diverse are the outputs compared to true survey responses?
    \item \textbf{\textit{Correlation structures:}} 
    Do the simulated responses reproduce the interdependencies between questions and topics that underpin the WVS and related social scientific research?   
\end{enumerate}

To outline the framework we use a motivating example similar to the experimental setup in similar value-alignment studies \citep{santurkar_whose_2023, durmus_towards_2024}. Here, we have:
\begin{itemize}[leftmargin=*, noitemsep, nolistsep] 
    \item a set of \textit{survey questions} $Q$ (e.g., the WVS); 
    \item a population with various \textit{demographic subgroups} $S$ (e.g., Germans, women, liberals, etc.) that has responded to each $q\in Q$; and
    \item  a set of \textit{models} $M$, where each $m\in M$ corresponds to a subgroup $s \in S$ and is used simulate responses to each $q\in Q$.
\end{itemize}

\subsection{Constructing and Aggregating Response Distributions}\label{sec:dist-agg-eval}

\begin{figure}[t!]
    \centering
    \includegraphics[width=\linewidth]{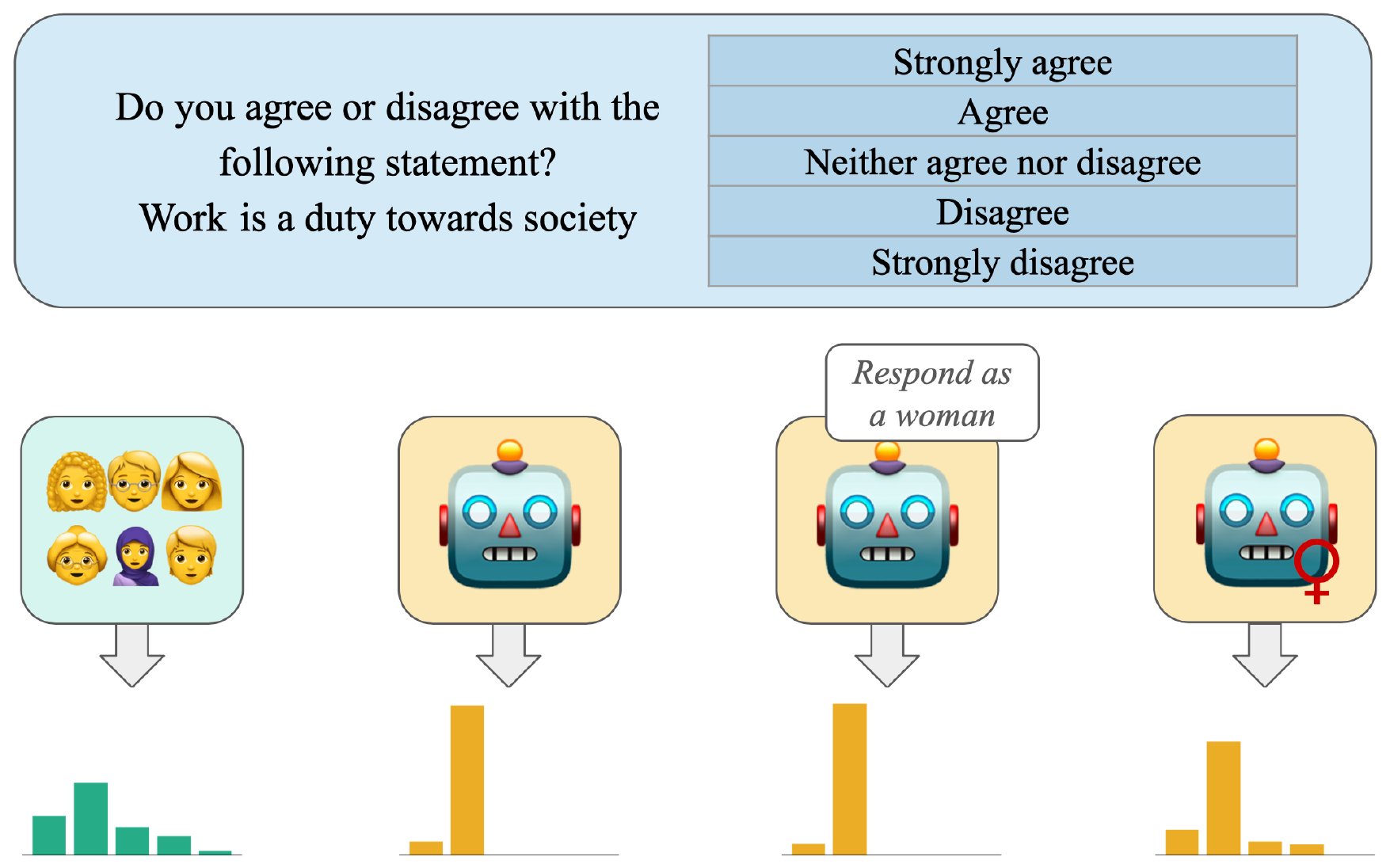}
    \caption{Process for constructing the marginal distributions of human opinions (green) and different steered models (yellow).}
    \label{fig:marginal}
\end{figure}

We first construct the \textbf{\textit{ground truth response distribution}} of each demographic subgroup $s \in S$ (for each question $q \in Q$) weighing responses using available survey weights. For each model $m \in M$ corresponding to the subgroup $s$, we then construct the matching \textbf{\textit{simulated response distribution}} using the generated outputs. This yields two distributions (one observed, $P_s$, and one simulated, $P_m$) of responses to each question from each  subgroup. Figure~\ref{fig:marginal} represents the construction of these marginal response distributions for human and simulated respondents.

To evaluate different groupings, responses can additionally be aggregated along one of the two axes before constructing the distributions: \begin{enumerate*}[label=(\arabic*)]
    \item \textit{\textbf{demographic dimensions}}, e.g., to investigate larger subsets of the respondents; and
    \item \textit{\textbf{question topics}}, e.g., to analyse broader value domains
\end{enumerate*}. 

\subsection{Evaluating Marginal Response Distributions}\label{sec:marginal-eval}

A major quantity of interest is the similarity between the empirical (marginal) response distribution, $P_s$, and the simulated ones, $P_m$, of the corresponding demographic subgroup \citep{durmus_towards_2024, santurkar_whose_2023}. We measure this with the marginal \textbf{\textit{dissimilarity}} (on $Q$) as:
$$\mathcal{D}(P_m, P_s) = \frac{1}{|Q|}\sum_{q\in Q} d(P_m(\cdot \mid q), P_s(\cdot \mid q))$$
where $d(\cdot,\cdot)\in[0,1]$ is a distance on probability distributions. $\mathcal{D}(P_m, P_s)$ is then the mean of per-question distances, yielding a unit-free dissimilarity score between 0 (perfectly representative) and 1 (maximal divergence). 

As a noted weakness of LLM-generated responses \citep{santurkar_whose_2023, bisbee_synthetic_2024, durmus_towards_2024}, we isolate response diversity by comparing the means of the normalised per-question variances for the true and generated responses. The \textbf{\textit{mean response variance}} is given by:
$$\mathcal{V}_s\left(P\right)= \frac{1}{|Q|} \sum_{q\in Q} \frac{\mathrm{Var}_{r \sim P(\cdot \mid q)}(r)}{\mathrm{diam}(R_q)^2}$$
where $P(r \mid q)$ denotes the probability assigned to response $r \in R_q$ for question $q$ under distribution $P \in \{P_s, P_m\}$. 
The variance $\mathrm{Var}_{r \sim P(\cdot \mid q)}(r)$ is computed over the response values $r$ with respect to this distribution and normalised by the diameter, $\mathrm{diam}(R_q)$, of the response set for question $q$.

\subsection{Evaluating Correlation Structures}\label{sec:correlation-eval}

Beyond marginal distributions, representativeness also requires that the simulated data preserve the \textbf{\textit{correlation structure}} between questions present in human responses. E.g., if respondents that score highly on \textit{Q1: importance of family} and also score highly on \textit{Q45: respect for authority}, then those items exhibit a positive correlation, potentially reflecting an underlying traditionalist worldview.
\begin{figure}[t!]
    \centering
    \includegraphics[width=1.0\linewidth]{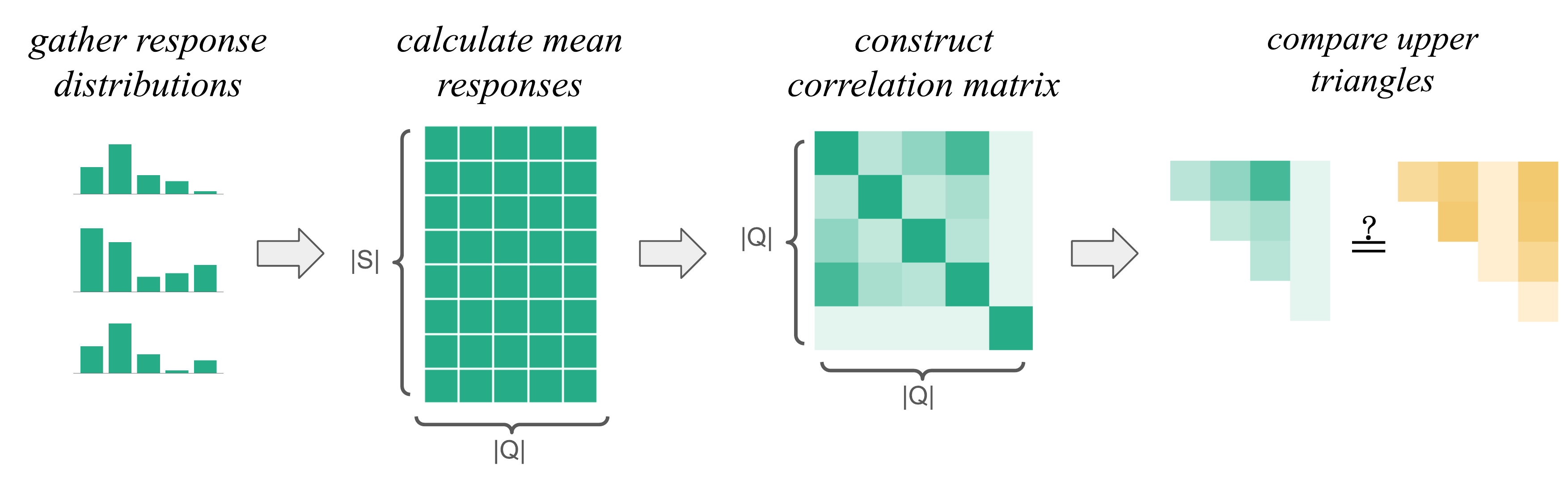}
    \caption{Process for constructing a correlation matrix from human opinions (green) and comparing it with a simulated correlation matrix (yellow).}
    \label{fig:correlation-process}
\end{figure}
As illustrated in Figure~\ref{fig:correlation-process}, we compare the correlation structures of empirical and simulated responses as follows:
\vspace{-2mm}
\begin{enumerate}[leftmargin=*, noitemsep]
    \item We gather the \textit{\textbf{response distributions}}, $P_s$ or $P_m$
    \item We compute the [0-1]-normalised \textit{\textbf{question mean response}} to item $q \in Q$ from subgroup $s \in S$ under each distribution, yielding two mean matrices $\mathrm{A}^\mathrm{true}$ and $\mathrm{A}^\mathrm{sim}$, both in $\mathbb{R}^{|S|\times |Q|}$
    \item We construct a \textbf{\textit{question-question correlation matrix}} $\mathrm{C}\in \mathbb{R}^{|Q|\times |Q|}$ as the pairwise correlation coefficients between columns\footnote{To construct a correlation matrix at a different level of aggregation (e.g., a \textit{topic-topic correlation matrix}), the same procedure applies with the extra step of first aggregating the response distributions accordingly.} of $A$ (for calculation details, see Appendix \ref{app:correlation})
    \item We compare the empirical correlation matrix $\mathrm{C}^{\text{true}}$ to the simulated one $\mathrm{C}^{\text{sim}}$ in two complementary ways: (1) the \textit{\textbf{Pearson correlation}}:  
$$
\rho_{\text{true}, \text{sim}} = \text{corr}\left(\mathrm{u}^{\text{true}}, \mathrm{u}^{\text{sim}} \right)
$$
as well as
(2) the \textit{\textbf{Root Mean Square Error (RMSE)}} (for the $n$ unique elements):
$$
\mathrm{RMSE}_{\text{true}, \text{sim}} = \sqrt{\frac{1}{n}\sum_{i=1}^n  \left(u^{\text{true}}_i - u^{\text{sim}}_{i} \right)^2}
$$

where $\mathrm{u}$ denotes the vector formed by stacking the upper-triangular, off-diagonal entries of a correlation matrix $\mathrm{C}$.
\end{enumerate}

This choice of two metrics\footnote{These metrics are used as a descriptive and comparative measure of structural alignment between vectorised correlation matrices rather than for statistical inference, thus, we do not rely on distributional assumptions.} (correlation and RMSE) allows us to separate whether a model reproduces the relative \emph{structure} of correlations (i.e., which question pairs tend to move together) from whether it also matches their \emph{magnitude}.

\section{Experimental Setup} 

Using the framework defined in \autoref{sec:framework}, we now compare persona prompting and fine-tuning as steering methods to make model outputs more representative of demographic subpopulations.

\subsection{Models}\label{sec:models}
We consider three model configurations: \begin{enumerate*}[label=(\arabic*)]
    \item an unsteered LLM;
    \item a persona prompted LLM; and 
    \item a demographic fine-tuned LLM: OpinionGPT
\end{enumerate*}. 
\paragraph{(1) Unsteered Baseline} \texttt{phi-3} \citep{abdin_phi-3_2024}. In this configuration, we do not try to steer the model to represent a specific demographic. This baseline allows us to assess the improvement of the following two steering methods. 
\paragraph{(2) Persona Prompting} \texttt{phi-3} + persona prompt. A simulation run with the addition of one of ten different demographic-specific profile instructions. This is the steered baseline and main point of comparison for demographic fine-tuning. 
\paragraph{(3) Demographic Fine-tuning}  We use an existing demographically fine-tuned model: OpinionGPT \citep{haller_opiniongpt_2024}, which consists of eleven fine-tuned LoRA adapters across four demographic variables (see \autoref{tab:opiniongpt-profiles} in the Appendix for an overview of adapters). OpinionGPT was created to make socio-demographic biases in LLMs explicit by training each adapter on data generated only by the target demographic subgroup. For this purpose, the authors leveraged data from the \textit{r/AskAXXX} subreddits, e.g., \textit{r/AskAGerman}. We do not include any additional steering in the prompt.

\vspace{2mm}
\noindent For each steered configuration, we generate responses for ten demographic subgroups: Two \textit{\textbf{gender}} attributes (\textit{male}, \textit{female}), two \textit{\textbf{age}}\footnote{We simulate responses for each demographic subgroup from OpinionGPT except for teenagers as this group is not represented in our evaluation data.} attributes (\textit{people over 30}, \textit{old people}), four \textit{\textbf{geographic}} attributes (\textit{German}, \textit{US American}, \textit{Latin American}, \textit{Middle Eastern}), and two \textbf{\textit{political}} attributes (\textit{liberal}, \textit{conservative}). Together with the unsteered baseline, we have 21 different models.

\subsection{Evaluation Data}
The World Values Survey (WVS) \citep{inglehart_globalization_2000} is an academic study of social, political, economic, religious, and cultural values and is conducted periodically with a new survey wave released every 5-10 years. The set of categories remains broadly consistent across waves to facilitate longitudinal and cross-national comparisons. Its multiple choice format allows for easy quantitative comparison, which combined with the complex and abstract nature of the survey topics makes it a popular resource for assessing LLM value alignment \citep{durmus_towards_2024, li_culturellm_2025}.
The close correspondence of the OpinionGPT modules to the data's demographic composition allows a like-for-like comparison for the available subgroups (see \autoref{tab:opiniongpt_wvs}).
\begin{table}[!h]
    \centering
    \begin{tabular}[\textwidth]{ll}
        \toprule
        OpinionGPT & WVS \\
        \midrule
        Liberal & \makecell[l]{respondent answered 1, 2, or 3 on a\\ 1-10 scale from political left to right} \\
        Conservative & \makecell[l]{respondent answered 8, 9 or 10 on a \\1-10 scale from political left to right} \\
        \hline
        German & respondent from Germany \\
        US American & respondent from the US \\
        Latin America & \makecell[l]{respondent from any available\\ Latin American country} \\
        Middle East & \makecell[l]{respondent from any available\\ Middle Eastern country} \\
        \hline
        Men & respondent is male \\
        Women & respondent is female \\
        \hline
        People Over 30 & \makecell[l]{respondents born after 1980 but\\ over 30 years of age} \\
        Old People & \makecell[l]{respondent born in or before 1980\\ (definition from subreddit)} \\
        \bottomrule
    \end{tabular}
    \caption{All OpinionGPT demographic subgroups from the political, geographic, gender, and age dimensions and their respective match in the WVS data. }
    \label{tab:opiniongpt_wvs}
\end{table}

We take survey data from WVS wave 7 \citep{haerpfer_world_2020}, which organises questions into various topics (see \autoref{app:wvs_data} for a summary). We select a subset of 193 questions that support easy comparison across the ten OpinionGPT subgroups and the use of a standard prompting structure like in similar studies \citep{santurkar_whose_2023} and QA tasks more generally \citep{liang_holistic_2023}. We use the human survey response distributions as a ground-truth to assess LLM representativeness.  \autoref{fig:marginal} also shows an example question.
\subsection{Prompting and Simulation}
For each of the $21$ models, we simulate $193$ questions with $500$ samples per question at a temperature of $0.9$ 
(see Appendix \ref{app:temp} for temperature analysis). After simulation we identify any responses that cannot be matched to an admissible response (e.g., due ambiguity or deviation from task) and mark this as invalid, analogous to a non-response in a human survey (see Appendix \ref{app:simulation} for details). 

We keep the system prompt simple with a brief description of the task and a single example to define the desired output format (see Appendix \ref{app:prompts}). Our persona prompting approach is similar to previous work \citep{bisbee_synthetic_2024, durmus_towards_2024, santurkar_whose_2023} where a short text is added to the prompts instructing the model to respond from the perspective of the relevant demographic subgroup. The format is shown in Appendix \ref{app:prompts}.

\subsection{Response Distributions and Aggregation}\label{sec:experiment-dists}
\begin{figure*}[tb!]
    \centering
    \includegraphics[width=0.495\textwidth]{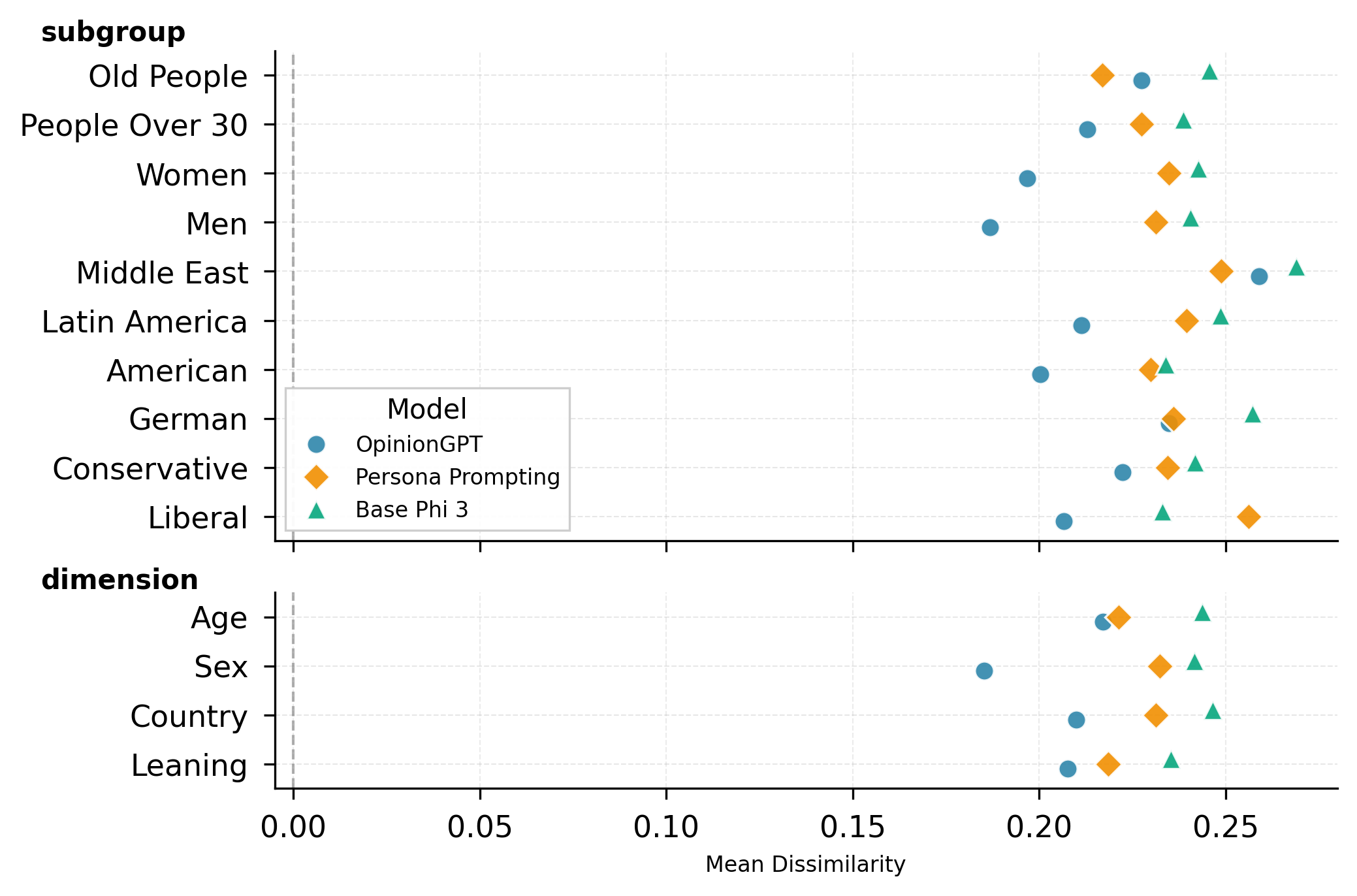}
    \includegraphics[width=0.495\textwidth]{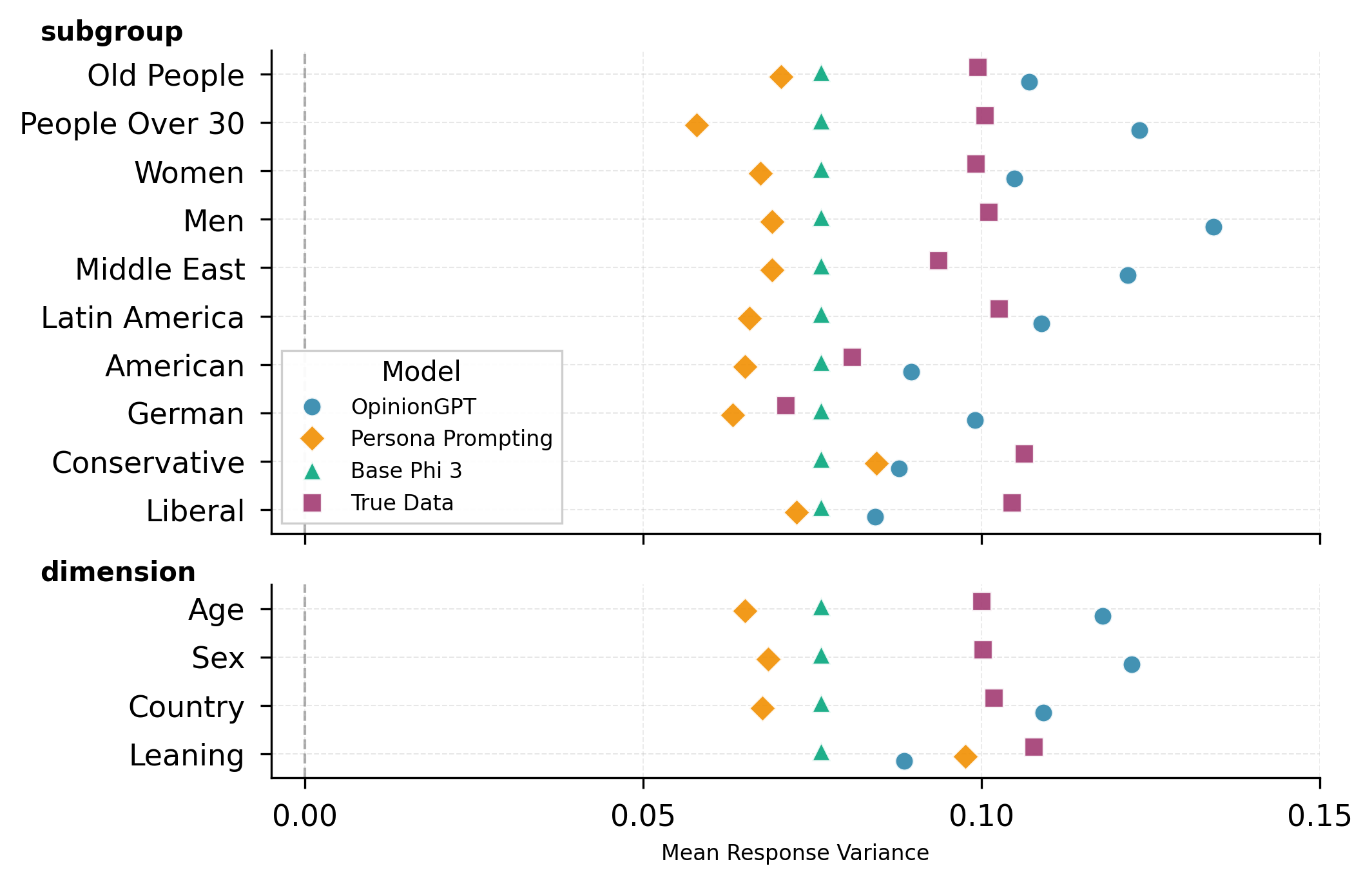}
    \caption{Evaluation metrics by demographic subgroup and dimension. Left: mean dissimilarity (lower = better). Right: mean response variance (closer to true data = better).}
    \label{fig:demographic-eval}
\end{figure*}

For each WVS question $q$, we have 10 subgroup-specific empirical distributions. We construct 21 model response distributions as outlined in Section~\ref{sec:models} (10 subgroups $\times$ 2 steering strategies plus unsteered baseline). We then aggregate once along each axis (demographics and questions). \textit{\textbf{Demographic dimension aggregation}} aggregates the data over all available profiles within a given dimension (e.g. \textit{men} and \textit{women} in the gender dimension) resulting in, for each question $q$, a distribution for gender, political leaning, geographic/cultural origin, and age (see: \autoref{tab:opiniongpt_wvs_details}); each can be seen as an approximation with a single response distribution of the entire WVS survey population using available demographic subgroups.

For \textbf{\textit{question topic aggregation}}, we aggregate all questions from the same thematic category of the WVS, resulting in 12 separate aggregated distributions for each subgroup $s$ or model $m$. This enables us to evaluate whether models capture the \textit{relationships between broad value domains} rather than just item-level associations.

\section{Analysis 1: Marginal Distributions}\label{sec:marginal-analysis}
This first analysis takes the traditional evaluation approach, investigating whether each model configuration can steer model outputs to better match the empirical marginal response distributions.

\subsection{Setup}

Using the response distributions from the \autoref{sec:experiment-dists}, we follow the procedure outlined in \autoref{sec:marginal-eval}. Here, we use the diameter-normalised \textit{Wasserstein-1} as our distance metric, $d$, for questions with an ordered response scale (e.g., Likert-type or numerical ratings), and the \textit{total variation} distance for  questions with a nominal response set. 
See Appendix \ref{app:metrics} for details.

\subsection{Results}
The left-hand panel in Figure~\ref{fig:demographic-eval} plots 
dissimilarity scores. It shows that model steering (through either OpinionGPT or persona prompting) produces more representative marginal response distributions than the unsteered baseline model. OpinionGPT reduces mean dissimilarity over the unsteered baseline for every subgroup, as does persona prompting for all but the \textit{liberal} demographic. The  improvement from OpinionGPT is greater than that from persona prompting for all but two demographic subgroups.

Similar patterns can be seen when aggregating along the demographic dimensions (Figure \ref{fig:demographic-eval}, bottom left). OpinionGPT and persona prompting improve noticeably over the baseline for each dimension, with large improvements for OpinionGPT compared to persona prompting across all dimensions. However, markedly different patterns emerge with question topic aggregation (see Appendix \autoref{fig:category-eval}). Both OpinionGPT and persona prompting reduce mean dissimilarity relative to the unsteered baseline, but the extent of improvement varies substantially across value domains. 
The right-hand panel of Figure~\ref{fig:demographic-eval} shows the mean response variance by demographic subgroup for each model configurations and the true responses. In each subgroup, OpinionGPT induces more response diversity to the responses than the unsteered baseline, whereas persona prompting decreases it. 
One reason for the suppressed variance is the collapse to a degenerate response distribution, which happens more frequently for persona prompting (see Appendix \ref{app:collapse}). An exception is the \textit{political leaning} subgroups and dimension where all model configurations suppress response diversity to below empirical levels. 

\section{Analysis 2: Correlation Structures}

Our first analysis found promising results in the models' ability to approximate marginal response distributions of the WVS. We now move beyond marginals to assess whether they can also produce the latent structures underpinning cultural theories \cite{hofstede_cultures_1980,schwartz_universals_1992}.

\subsection{Setup}

We analyse correlation matrices at two levels: \begin{enumerate*}[label=(\roman*)]
    \item the \textit{\textbf{question–question}} level, which considers correlations between individual survey questions\footnote{This calculation applies only to ordinal questions $Q^{ord}\subset Q$, omitting the $<$10\% nominal-answer questions.}; and
    \item the \textit{\textbf{topic–topic}} level, where questions are collapsed into the 12 thematic value domains in our WVS subset 
\end{enumerate*}. The first tests whether models capture fine-grained associations between specific attitudes, while the second evaluates whether they reproduce broad inter-domain dependencies.

We construct and compare \textit{question-question correlation matrices} as in \autoref{sec:correlation-eval}. To construct the \textit{topic-topic correlation matrix}, the same procedure applies with the extra step of first aggregating responses to the thematic domains. To account for sampling variability in the simulated responses, we report bootstrapped 95\% confidence intervals for each metric (RMSE and Pearson $\rho$) computed from 500 resamples (with replacement) of the 500 generated responses for each survey item from each model. For full calculation details of the matrices and bootstrap, see Appendix \ref{app:eval}. As we use question means in this experiment to construct the correlation matrix, we leave out the $18$ categorical questions, which lack the notion of a mean. This omits $<$10\% of the $193$ questions, more details in Appendix~\ref{app:categorical-omission}.

We further contextualise the proposed correlation-structure metrics by estimating lower and upper bounds on model performance via a \textit{\textbf{permutation-based null}} baseline and a \textit{\textbf{split-half resampling}} of the WVS, respectively. 

\textit{The permutation-based null} provides a baseline\footnote{Our baseline model is demographic-agnostic, this lack of this axis means there are is no subgroup correlation structure. The permutation-based null serves as a baseline instead.} corresponding to the absence of cross-question correlation between subgroups. We construct this by independently permuting the mean WVS responses (each column of $\mathrm{A}^{\mathrm{WVS}}$) before recalculating the correlation matrix and comparing this to the true $\mathrm{C}^{\mathrm{WVS}}$ using the same evaluation metrics. This preserves the marginal distribution of response means for each survey item while destroying correlations between items. 

The \textit{split-half resampling} complements this by providing an empirical ceiling for the metrics through an estimate for the most optimistic case, i.e., when the correlation matrices are constructed from subsets of the exact same distribution of responses. To perform this, we randomly partition WVS responses into two halves, computing correlation matrices for each subset and again comparing them using the same evaluation metrics. Each is performed $1,000$ times with the percentile values reported corresponding to a 95\% interval. Details can be found in Appendix \ref{app:correlation-lb-ub}.

\subsection{Results}
Relative to a structureless permutation-based null baseline and a near-noiseless split-half resampling, the steering methods do recover a degree of the inter-subgroup correlation structure at the question-question level (see \autoref{tab:question-correlation_metrics}). However, both OpinionGPT and persona prompting still remain well below the empirical ceiling, indicating poor recovery of fine-grained dependence patterns. While neither model accurately reproduces the \textit{relative pattern} of correlations, persona prompting does so noticeably better than OpinionGPT ($\rho$: $0.158$ vs $0.09$). The respective bootstrapped intervals (see \autoref{tab:question-correlation_metrics}) are tight around the point estimates, indicating that the effect of sampling variability of these metrics is fairly low. Despite displaying stronger directional similarity than OpinionGPT, from the RMSE we can see that persona prompting produces correlation patterns that are further away in \textit{magnitude} (RMSE: $0.679$ vs $0.638$). Again, the tight confidence intervals seen in \autoref{tab:question-correlation_metrics} are tight around the point estimates.

\begin{table}[!tbp]
\centering
\small
\begin{tabular}{lS[table-format=1.3, table-space-text-post={[ -0.00, 0.00 ]}]
S[table-format=1.3, table-space-text-post={[ 0.00, 0.00 ]}]}
\toprule
Model & \multicolumn{1}{l}{Pearson $\rho$} & \multicolumn{1}{l}{RMSE} \\
\midrule
OpinionGPT & 0.090 [0.08, 0.10] & 0.638 [0.63, 0.64] \\
Persona & 0.158 [0.15, 0.17] & 0.679 [0.67, 0.68] \\
Perm. Null & -0.004 & 0.849 \\
Split Half & 0.999 & 0.006 \\
\bottomrule
\end{tabular}

\caption{Question-question correlation metrics with 95\% bootstrapped confidence intervals in brackets}
\label{tab:question-correlation_metrics}
\end{table}

When aggregating items into question topics, the preservation of correlation structures from OpinionGPT degrades notably (see \autoref{tab:category-correlation_metrics}). With $\rho$: $-0.018$ the relative pattern of correlations is now completely lost; the RMSE is also higher in comparison to the question-question case. Conversely, persona prompting improves on both metrics after aggregation ($\rho$: $0.24$, RMSE: $0.676$). Once again, the confidence intervals are narrow around the point estimates. Although the results in \autoref{sec:marginal-analysis} show that OpinionGPT produces marginal distributions that are much closer to the WVS ground truth, the results of this analysis show that persona prompting performs marginally better at reproducing the empirical correlation structure across subgroups. Nevertheless, neither technique faithfully reproduces these correlation patterns.

\begin{table}[!tbp]
\centering
\small
\begin{tabular}{lS[table-format=1.3, table-space-text-post={[ -0.00, 0.00 ]}]
S[table-format=1.3, table-space-text-post={[ 0.00, 0.00 ]}]}
\toprule
Model & \multicolumn{1}{l}{Pearson $\rho$} & \multicolumn{1}{l}{RMSE} \\
\midrule
OpinionGPT & -0.018 [-0.02, 0.05] & 0.718 [0.71, 0.73]\\
Persona & 0.240 [0.21, 0.28] & 0.676 [0.67, 0.69]\\
Perm. Null & 0.001 & 0.914\\
Split Half & 0.997 & 0.011\\
\bottomrule
\end{tabular}

\caption{Topic-topic correlation metrics with 95\% bootstrapped confidence intervals in brackets}
\label{tab:category-correlation_metrics}
\end{table}

\section{Discussion}

\paragraph{Representativeness of Marginal Distributions.}
We find mixed results for our \ref{rq1}: OpinionGPT improved marginal distribution similarity over the unsteered baseline across all demographics (see \autoref{fig:demographic-eval}). Improvements were also partially evident for persona prompting, in line with previous findings \citep{bisbee_synthetic_2024, santurkar_whose_2023}. 
\citep{leidinger_language_2023} or algorithmic search \cite{zheng_when_2024}. 
Yet, while the relative improvements were far more consistent and substantial for OpinionGPT, distributional dissimilarity is not eliminated and remains uneven and context-dependent; subgroups such as \textit{Middle East} and \textit{old people} are less well represented, potentially due to poor representation in the Reddit fine-tuning data. 

Response diversity has been noted as a particular weakness with LLM responses \citep{santurkar_whose_2023, bisbee_synthetic_2024, durmus_towards_2024}. As seen in \autoref{fig:demographic-eval}, OpinionGPT induces greater response diversity compared to the unsteered baseline, which in line with previous findings is already less variant than true responses. However, the variance of OpinionGPT responses often exceeds empirical levels, producing distributions that overstate the degree of disagreement within a demographic group. Persona prompting displays the opposite problem and greatly suppresses response diversity. The mechanisms of the respective steering methods differ substantially. Fine-tuning reshapes model parameters using data directly from the target demographic and in doing so appears to induce more stable and heterogeneous marginal response distributions. In contrast, persona prompting functions as a strong conditioning signal that narrows the effective response space. This signal often collapses the distribution toward a single stereotypical response (see Appendix \autoref{tab:degenerate-counts} for more detail), completely suppressing the heterogeneity observed in the WVS; this is a commonly highlighted issue with persona prompted models \citep{durmus_towards_2024, gupta_bias_2024}.

\paragraph{Correlation Structures and Aggregation Effects.} 
 
Previous findings suggest that steered or aligned LLMs are better at capturing surface-level alignment patterns than deep latent structure \citep{santurkar_whose_2023, durmus_towards_2024, gupta_bias_2024, kabir_testing_2025}. Using our framework to answer \ref{rq2}, we add further evidence to this, emphasising that improved marginal distribution similarity through model steering does not imply the preservation of latent structures. While persona prompting performs slightly better than OpinionGPT, neither model is able to adequately able to capture question-question correlation structure of WVS data (\autoref{tab:question-correlation_metrics}) or the topic–topic correlation (\autoref{tab:category-correlation_metrics}). This highlights a key weakness: Simulated responses can approximate marginal response distributions to an extent but have more difficulty reproducing the higher-order relationships between value domains that constitute a coherent political worldview \citep{ceron_beyond_2024} and, which give surveys such as the WVS their interpretive coherence. Again, the differing steering mechanisms may offer one possible explanation. OpinionGPT uses separate adapters for each demographic group, which may allow representations to drift across groups. In contrast, persona prompting relies on a single conditioned model, meaning that responses across groups are generated from a shared underlying representation. The greater heterogeneity previously observed for OpinionGPT may therefore help reproduce diverse marginal response distributions, but could be disadvantageous when modelling the relational structure across demographic subgroups.

Finally, \ref{rq3} examines whether the representativeness observed in simulated responses persists when results are aggregated. The results suggest that aggregation by demographic dimension does not substantially diminish representativeness, preserving marginal similarity and variance patterns. In contrast, aggregation by question topic produced less stable results, with marginal similarity varying substantially across domains, in addition to the above mentioned issues with correlation structure preservation. This suggests that representativeness depends not only on demographic context, but also on the question topic. Despite some limitations, OpinionGPT better approximates marginal response distributions than the unsteered baseline across value domains and, with the exception of \textit{Ethical Values}, outperforms or roughly equals persona prompting. However, after the same aggregation to topic level the correlation structure of OpinionGPT responses loses all relative similarity to the true WVS, whereas this slightly improves for persona prompting, despite remaining poor overall.

\paragraph{Broader Implications for Representativeness.}

The findings underscore that representativeness constitutes a distinct axis of model alignment, separate from established dimensions such as safety, helpfulness, or factuality \citep{gabriel_artificial_2020}. These can often be assessed at the level of individual responses, but representativeness is inherently defined at the distributional level, requiring the preservation not only of central tendencies but also of variance, demographic fidelity, and correlation structures \citep{santurkar_whose_2023, argyle_out_2023, dominguez-olmedo_questioning_2024}. Our results demonstrate that our tested approaches lack this representativeness. Therefore, long-standing cultural values theories such as those of Hofstede and Schwartz, which show that meaningful cultural differences emerge from latent structures derived from multivariate correlation in survey data rather than from any single attitude in isolation, do not hold for response from our tested LLMs. The results from our empirical case study highlight the importance of evaluating both marginal response distributions and inter-item correlations when assessing demographic alignment. Our framework enables this joint evaluation and reveals differences between steering approaches that would be overlooked by analyses focusing on marginals alone.

\section{Conclusion}

In this paper, we have argued that representativeness is a distinct and necessary dimension of alignment, one that requires the preservation of not only diversity and subgroup fidelity but also the structural interdependencies that characterise human populations and underpin major cultural values frameworks in the social sciences. We therefore propose an evaluation framework that explicitly considers both marginal response distributions and inter-question correlation structures. Applying this to the WVS, we tested two model steering approaches for demographic alignment: \begin{enumerate*}[label=(\arabic*)]
    \item demographic fine-tuning, with OpinionGPT; and
    \item persona prompting, with Phi-3-Mini-Instruct-4k
\end{enumerate*}. Our results show that while OpinionGPT shows promising results in marginal similarity and variance structure when compared to persona prompting, it performs worse at reproducing the correlation structures, although neither steering approach was able to faithfully preserve these higher order interdependencies. We thus emphasise the fragility of demographic alignment and the danger of making representativeness claims based on marginal distribution properties alone.

An important direction for future work is the incorporation of population-level dependency structures into alignment mechanisms to enable both steerability across groups as well as fidelity to population-wide distributions, especially regarding the interconnectedness of how values and opinions are structured.

\section*{Limitations}

\paragraph{Model limitations.} OpinionGPT's base model, \textit{Phi 3 Mini 4k}, is a compact LLM with far fewer parameters and a shorter context window than frontier-scale models. Through targeted data curation, the authors nonetheless achieve strong performance, often comparable to larger models \citep{gunasekar_textbooks_2023, abdin_phi-3_2024}, but also note size-related limits that may reduce the ability to capture complex demographic variation. Smaller models are more prone to deviating from instructions and prescribed response formats, adding noise and errors \citep{murthy_evaluating_2024}. OpinionGPT was also fine-tuned on Reddit data, whose users skew young \citep{pew_research_center_social_2024} and whose lingua franca is English rather than region-specific languages, limiting representativeness via the fine-tuning corpora. Finally, focusing on a single model is itself a limitation. Our intent is to present a nuanced evaluation framework, which future work could extend to other alignment methods and base models.

\paragraph{Design choices.} We independently sample 500 responses per item (for each model configuration) and compare distributions. While common in simulated survey alignment work \citep{santurkar_whose_2023, durmus_towards_2024}, this requires aggregating by demographic subgroup to construct correlation matrices (see \autoref{app:correlation}), losing some distributional information. A trajectory-based alternative would condition each answer on prior ones to generate full respondent surveys from which correlation matrices could be constructed. This would allow respondent-level analyses and finer-grained item correlations while remaining compatible with our framework. In \autoref{sec:correlation-eval}, one would skip mean-response computation in step 2 and construct correlations in step 3 directly from trajectories.

\paragraph{WVS and multiple choice surveys.} The WVS, rooted in a European values project, embeds Western and Eurocentric normative assumptions \citep{goodwin_cross-cultural_2020}, shaping both question design and interpretation and making representativeness relative to a culturally contingent benchmark rather than a neutral truth. Our analysis is also English-only, so we cannot claim generalisability to other languages. For geographic subgroups, prompting in region-native languages (e.g., Arabic for the Middle East) might yield better alignment than English, given likely regional differences in pre-training data sources. Multiple-choice surveys further offer clear advantages such as straightforward encoding, comparable probability distributions, and large-scale robust resources, hence their use in survey research and LLM alignment \citep{santurkar_whose_2023, durmus_towards_2024}. However, the closed-ended tasks necessarily restrict nuance and may miss misalignment that appears in open-ended settings.

\paragraph{Demographic Subgroups.} We evaluate ten demographic subgroups defined by OpinionGPT adapters, which do not exhaust relevant respondent characteristics. Coverage is limited to male/female genders, two US-spectrum political views, and no people below 30, while geographic groups differ in granularity (single-country groups like Germans or US Americans versus broader regions like Middle Easterners or Latin Americans spanning diverse contexts). Other dimensions (e.g., socio-economic status or education) may also matter, and our one-dimensional subgrouping ignores intersectional perspectives.

\section*{Ethics statement}
All our data is publicly available and licensed for research. We did not fine-tune any models ourselves, but the models we use have been fine-tuned on publicly available data only. The WVS includes real data from survey respondents, but they have been anonymised before the publication of the WVS data. We also caution against conflating high scores on specific representativeness metrics with genuine model alignment.

\section*{Acknowledgements}
Alan Akbik is supported by the Deutsche Forschungsgemeinschaft (DFG, German Research Foundation) under Emmy Noether grant "Eidetic Representations of Natural Language" (project number 448414230). Further, Alan Akbik is supported by the Deutsche Forschungsgemeinschaft (DFG, German Research Foundation) under Germany’s Excellence Strategy "Science of Intelligence" (EXC 2002/1, project number 390523135).
\bibliography{acl-refs}

\appendix

\section{OpinionGPT}

OpinionGPT is a demographically fine-tuned model \citep{haller_opiniongpt_2024} that aims to make biases explicit and transparent rather than attempting to eliminate or obscure them. Low rank adaptation (LoRA) to fine-tune a set of modules, each corresponding to a particular demographic profile enabling users to generate responses from the perspective of each of these groups. In this way, OpinionGPT facilitates controlled exploration of how demographic factors may influence language model outputs, thereby attempting to address the issue of hidden or unacknowledged biases. To construct the demographic LoRA modules, the authors identified a set of subreddits from the \textit{AskX} schema, each associated with a particular demographic subgroup. The full list of demographic modules, their corresponding subreddits, and the sample sizes is provided in \autoref{tab:opiniongpt-profiles}.

\begin{table}[ht]
    \centering
    \begin{tabular}{llr}
        \toprule
        \textit{Demographic} & \textit{Source Subreddit} & \textit{Sample Size} \\
        \midrule
        \multicolumn{3}{c}{\textbf{Geographical}} \\
        \addlinespace
        German         & AskAGerman        & 11k \\
        American       & AskAnAmerican     & 20k \\
        Latin American & AskLatinAmerica   & 20k \\
        Middle East    & AskMiddleEast     & 20k \\
        \midrule
        \multicolumn{3}{c}{\textbf{Political}} \\
        \addlinespace
        Liberal        & AskALiberal       & 20k \\
        Conservative   & AskConservatives  & 18k \\
        \midrule
        \multicolumn{3}{c}{\textbf{Gender}} \\
        \addlinespace
        Female         & AskWomen          & 20k \\
        Male           & AskMen            & 20k \\
        \midrule
        \multicolumn{3}{c}{\textbf{Age Demographics}} \\
        \addlinespace
        Teenager (girls)     & AskTeenGirls      & 10k \\
        Teenager (boys)      & AskTeenBoys       & 10k \\
        Over 30 (men)        & AskMenOver30      & 10k \\
        Over 30 (women)      & AskWomenOver30    & 10k \\
        Old people           & AskOldPeople      & 15.5k \\
        \bottomrule
    \end{tabular}
    \caption{List of all demographic LoRA modules and their corresponding subreddits \citep{haller_opiniongpt_2024}}
    \label{tab:opiniongpt-profiles}
\end{table}

As briefly described in Table \ref{tab:opiniongpt_wvs_details}, we have eleven OpinionGPT adapters, of which we consider ten for this analysis. We exclude the teenagers subgroup since there is no good match in the WVS data: While teenagers range from ages 13 to 19, the WVS has respondents age 18 and older only. For all other adapters, we identify the relevant sociodemographic questions and filter all responses based on these questions. Table \ref{tab:opiniongpt_wvs_details} shows all OpinionGPT subgroups, their corresponding reddit channels and additional channel definitions, and the WVS questions and values we used for filtering. 

\begin{table*}[]
    \centering
\begin{tabular}{lllll}
\toprule
\begin{tabular}[c]{@{}l@{}}Demographic \\ Dimension\end{tabular} &
  \begin{tabular}[c]{@{}l@{}}OpinionGPT \\ Subgroup\end{tabular} &
  Reddit Channel &
  WVS Question &
  \begin{tabular}[c]{@{}l@{}}WVS Question \\ Values for Subgroup\end{tabular} \\
  \midrule
\multirow{2}{*}{Political} &
  Liberal &
  AskALiberal &
  \multirow{2}{*}{\begin{tabular}[c]{@{}l@{}}Q240: \\ In political matters, \\ people talk of "the \\ left" and "the right." \\ How would you \\ place your views \\ on this scale, \\ generally speaking?\\ 1 (left) to 10 (right)\end{tabular}} &
  1-3 left \\
 \addlinespace[9ex]
 &
  Conservative &
  AskConservatives &
   &
  8-10 right \\
 \addlinespace[9ex]
  \midrule
\multirow{4}{*}{Geographic} &
  German &
  AskAGerman &
  \multirow{4}{*}{\begin{tabular}[c]{@{}l@{}}Q266\\ In which country \\ were you born?\end{tabular}} &
  Germany \\
 &
  America &
  AskAnAmerican &
   &
  US \\
 &
  Latin American &
  \begin{tabular}[c]{@{}l@{}}AskLatinAmerica\\ (Latin America \\ and the Caribbean)\end{tabular} &
   &
  \begin{tabular}[c]{@{}l@{}}Argentina, Bolivia, Brazil, \\ Chile, Colombia, Ecuador, \\ Guatemala, Mexico, \\ Nicaragua, Peru, Puerto Rico, \\ Uruguay, Venezuela\end{tabular} \\
 &
  Middle Eastern &
  \begin{tabular}[c]{@{}l@{}}AskMiddleEast\\ (Middle East and \\ North Africa)\end{tabular} &
   &
  \begin{tabular}[c]{@{}l@{}}Cyprus, Egypt, Iran, Iraq, \\ Jordan, Lebanon, Turkey\end{tabular} \\
  \midrule
\multirow{2}{*}{Gender} &
  Men &
  AskMen &
  \multirow{2}{*}{\begin{tabular}[c]{@{}l@{}}Q260: \\ Respondent's sex\end{tabular}} &
  1 male \\
 &
  Women &
  AskWomen &
   &
  2 female \\
  \midrule
\multirow{3}{*}{Age} &
  Teenagers &
  \begin{tabular}[c]{@{}l@{}}AskTeenGirls / \\ AskTeenBoys\end{tabular} &
  \multirow{3}{*}{\begin{tabular}[c]{@{}l@{}}Q261: \\ Can you tell me \\ your year of birth?\\ Q262: \\ This means you \\ are XX years old?\end{tabular}} &
  \begin{tabular}[c]{@{}l@{}}N/A, excluded because WVS \\ contains ages 18+ only\end{tabular} \\
 &
  People over 30 &
  \begin{tabular}[c]{@{}l@{}}AskWomenOver30 / \\ AskMenOver30\end{tabular} &
   &
  \begin{tabular}[c]{@{}l@{}}born after 1980, \\ age over 30\end{tabular} \\
 &
  Old People &
  \begin{tabular}[c]{@{}l@{}}AskOldPeople\\ (born in or\\ before 1980)\end{tabular} &
   &
  born in or before 1980\\
  \bottomrule
    \end{tabular}
    \caption{All OpinionGPT demographic subgroups from the political, geographic, gender, and age dimensions, the wording of the filter question, and which values defined the subgroup.}
    \label{tab:opiniongpt_wvs_details}
\end{table*}
On the political dimension, reddit channels target Liberals and Conservatives, which might be more targeted to the US context, while the question we use operates on a left/right dimension, which is more universal. We still only use responses that show a clear left or right leaning stance.

For the geographic adapters, we used the definitions of the Middle East and Latin America from Wikipedia\footnote{\url{https://en.wikipedia.org/wiki/Middle_East}, \url{https://en.wikipedia.org/wiki/Latin_America}}. While \textit{America} could refer to multiple countries, the channel logo (Uncle Sam with elements of the US flag) clearly refers to the US only. 

We also define the age groups to be distinct, i.e., we use the definition of \textit{old people} from the reddit channel and we define \textit{people over 30} to be at least 30, but younger than the \textit{old people} group.  

\section{Data, Simulation and Prompting}\label{app:simulation}

\subsection{WVS Data Subset}\label{app:wvs_data}

We select a subset of WVS questions to enable easy comparison and a consistent prompting approach. We consider a question "easily comparable" if it \begin{enumerate*}[label=(\alph*)]
    \item is not country-specific, allowing comparison across all geographic subgroups; and
    \item does not depend on other questions, enabling an easy prompting structure (as used in similar works \citep{santurkar_whose_2023} as well as for QA tasks more generally \citep{liang_holistic_2023}) without the need to reword the question 
\end{enumerate*}. A summary of included questions by topic can be seen in \autoref{tab:wvs-summary}.

\begin{table}[!hbtp]
    \centering
    \begin{tabular}{lll}
\toprule
Topic & \makecell[l]{Count In \\Complete \\Survey} & \makecell[l]{Count In \\Our Subset} \\
\midrule
\makecell[l]{Social values, \\norms, stereotypes} & 45 & 22 \\
\makecell[l]{Economic values} & 6 & 6 \\
\makecell[l]{Perceptions of corruption} & 9 & 9 \\
\makecell[l]{Perceptions of migration} & 10 & 10 \\
\makecell[l]{Perceptions of security} & 21 & 21 \\
\makecell[l]{Index of postmaterialism} & 6 & 0 \\
\makecell[l]{Perceptions about \\science and technology} & 6 & 6 \\
\makecell[l]{Religious values} & 12 & 11 \\
\makecell[l]{Ethical values} & 23 & 22 \\
\makecell[l]{Political interest and \\political participation} & 36 & 19 \\
\makecell[l]{Political culture and \\political regimes} & 25 & 25 \\
\makecell[l]{Happiness and wellbeing} & 11 & 11 \\
\makecell[l]{Social capital, trust\\ and organizational \\membership} & 49 & 31 \\
\bottomrule
\end{tabular}

    \caption{Summary of included WVS questions by topic}
    \label{tab:wvs-summary}
\end{table}

\subsection{Simulation Procedure}

We use the following process to extract valid outputs and create response distributions for each question from each simulated subgroup.

\begin{itemize}[leftmargin=*, noitemsep]
    \item For each model (21 in total), we \textit{\textbf{initialise a new instance}} 
    \item We then \textbf{\textit{present each survey item}} (193 in total) to the model as a stand-alone prompt
    \item For each question (with each model) we \textit{\textbf{sample 500 responses}} (temperature $0.9$)
    \item We post-process raw generations to adhere to \textbf{\textit{desired response format}}. This involves stripping extraneous text, mapping the output to the corresponding numeric response option, and discarding invalid generations that could not be reliably matched
    \item Finally, we aggregate cleaned responses to \textbf{\textit{form response distributions}} for each question and subgroup
\end{itemize}

\subsection{Prompts}\label{app:prompts}

The system prompt used for all simulations is given in \autoref{fig:systemprompt}. For persona prompting, each subgroup was instantiated using a single natural-language persona description from the template in \autoref{fig:persona_template} using the descriptions in \autoref{tab:persona_descriptions}. Each persona prompt was then appended to the system prompt for the corresponding model. While persona prompting is widely used to condition LLM responses on demographic attributes, the outputs of large language models are known to be sensitive to prompt wording. At present, there is no clear consensus on how to systematically determine optimal prompts, either through linguistic analysis of prompt formulations \citep{leidinger_language_2023} or through automated search methods \citep{zheng_when_2024}. Consequently, we employ a fixed prompt template and subgroup descriptions to ensure reproducibility.

\begin{figure}
\begin{tcolorbox}[
    colback=white,
    colframe=black,
    title=System Prompt, 
    center title, 
    fontupper=\ttfamily,
    fonttitle=\ttfamily
]

You are taking part in a survey on personal values and will be asked a range of multiple choice questions. \\
Each question will have the following format:\\
\\
"\\
Q33: For each of the following statements I read out,
can you tell me how much you agree with each. Do you 
agree strongly, agree, disagree, or disagree strongly? - 
Carrot cake is tastier than chocolate cake
\\
Available responses:\\
1: Agree strongly\\
2: Agree\\
3: Disagree\\
4: Strongly disagree\\
"\\
\\
For each question, just select a single answer from the set of available responses. Please copy the answer exactly as written in the 'Available responses' list above. Do not change spelling or capitalisation. \\
For example:\\
4: Agree strongly

\end{tcolorbox}
\caption{System prompt used for all simulations}
\label{fig:systemprompt}
\end{figure}

\begin{figure}[!htbp]
\begin{tcolorbox}[
    colback=white,
    colframe=black,
    title=Persona Prompt Template, 
    center title, 
    fontupper=\ttfamily,
    fonttitle=\ttfamily
]

Please answer all questions as if you are \{\textit{subgroup\_description}\}. Make sure your responses align as closely as possible to the responses that someone from this demographic is likely to give.

\end{tcolorbox}
\caption{Template for the persona prompts.}
\label{fig:persona_template}
\end{figure}

\begin{table}[!htbp]
\centering
\begin{tabular}{lp{5cm}}
\toprule
\textbf{Subgroup} & \textbf{Persona Description} \\
\midrule
American & a person born and raised in the United States of America \\
German & a person born and raised in the Federal Republic of Germany \\
Latin America & a person born and raised in a Latin American country \\
Middle East & a person born and raised in a Middle Eastern country \\
Men & a person that identifies as male \\
Women & a person that identifies as female \\
People over 30 & a person aged between 30 and 45 \\
Old people & a person aged over 45 \\
Conservative & a person with a conservative or politically right-wing view of the world \\
Liberal & a person with a progressive or politically left-wing view of the world \\
\bottomrule
\end{tabular}
\caption{Subgroup descriptions used to instantiate the persona prompt template.}
\label{tab:persona_descriptions}
\end{table}

\newpage
\section{Evaluation}\label{app:eval}
\subsection{Response distributions}\label{app:dists}
We construct response distributions for a given set of questions $Q$, subgroups $S$ and, models $M$ as follows:
\begin{itemize}[leftmargin=*]
    \setlength\itemsep{.1em}
    \item For each \textit{subgroup} $s \in S$, the survey-weighted \textbf{\textit{ground truth response distribution}} for a question $q\in Q$ with possible responses $r\in R_q$ is given by:$$
    P_s(r \mid q) \;=\; 
    \frac{\sum_{i=1}^{n_{s \mid q}} w_i \,\mathbf{1}\{\,x^{\mathrm{WVS}}_{i \mid q, s} = r\,\}}
    {\sum_{i=1}^{n_{s \mid q}} w_i}
    $$
    where:
    \begin{itemize}[leftmargin=*, noitemsep, nolistsep]
        \item $x^{\mathrm{WVS}}_{i \mid q, s}$ denotes the response to question $q$ from a WVS respondent $x_i\in X^\mathrm{WVS}$ that belongs to subgroup $s\in S$
        \item $n_{s \mid q}$ is the number of observed WVS responses for item $q$ from subgroup $s$; and
        \item $w_i$ denotes the survey weight assigned to respondent $x_i\in X^\mathrm{WVS}$.
    \end{itemize}
     
    \item For each \textit{model} $m \in M$, the \textit{\textbf{simulated response distribution}} for a question $q\in Q$ with possible responses $r\in R_q$ is given by:
    $$
    P_m(r \mid q) = \frac{1}{n_{m \mid q}} \sum_{i=1}^{n_{m \mid q}} \mathbf{1}\{x^{m}_{i \mid q} = r\}
    $$
    where:
    \begin{itemize}[leftmargin=*, noitemsep, nolistsep]
        \item $x^{m}_{i \mid q}$ denotes the simulated response to question $q$ a model $m$; and
        \item $n_{m \mid q}$ is the number of samples drawn from model $m$ for question $q$.
    \end{itemize}
\end{itemize}

\subsection{Result Aggregation}\label{app:aggregation}

To evaluate the robustness of the models, we compare results not only by question for each model configuration, but additionally by aggregating along two different axes corresponding to \begin{enumerate*}[label=(\arabic*)]
    \item the \textit{demographics} (and by extension the models); and
    \item the \textit{survey questions}
\end{enumerate*}.

The two levels of \textbf{\textit{demographic aggregation}} align with the OpinionGPT modules. \textit{Demographic subgroup aggregation} aggregates responses over each of the subgroups corresponding to the OpinionGPT modules before calculating an evaluation metric. This isolates each separate simulation run, focusing on the effectiveness of the individual models. \textit{Demographic dimension aggregation} groups the data across all available subgroups in a given dimension (see Appendix \autoref{tab:opiniongpt-profiles} for list of demographic subgroups and dimensions) before metric calculation; this can be seen as an approximation to the entire WVS survey population.

For \textbf{\textit{survey level aggregation}} we collapse survey items into the thirteen thematic categories of the WVS. In addition to the \textit{individual question}, where each question's response set for each question are analysed separately, providing the \textit{most fine-grained view} of model alignment and capturing local covariances between questions, we also aggregate along \textit{question topic}, where item responses are aggregated within the questions categories across all demographic subgroups. Analyses at this level enable us to evaluate whether models capture the \textit{relationships between broad value domains} rather than only item-level associations, and therefore whether representativeness extends to the higher-order structures that underpin social scientific cultural theories.

\subsection{Metrics for Marginal Similarity}\label{app:metrics}

\paragraph{Wasserstein distance} measures the dissimilarity between two probability distributions by measuring the minimal cost of reconfiguring the mass of one distribution such that you recover the other \citep{villani_wasserstein_2009}. Unlike divergence metrics such as Jensen-Shannon, it takes order into account making it particularly suited to the Likert-type questions common in the WVS. However, it is not suitable for non-ordinal distributions and the unmodified score is not bounded and therefore might be more difficult to interpret.

In the one-dimensional, discrete case the Wasserstein distance (for each $q\in Q^\mathrm{ord}$) is given by:
$$W(P_m, P_s)=\sum_{r \in R_q} \|F_m(r)-F_s(r)\|$$
where $F_m$ and $F_s$ are the corresponding empirical CDFs to $P_m$ and $P_s$.For improved interpretability and consistent comparison between questions (and their differing response sets), we use the following normalised version of the Wasserstein distance (similar to that in \citet{santurkar_whose_2023}):
$$d_{\mathrm{W}}(P_m, P_s) := \frac{W(P_m, P_s)}{\mathrm{diam}(R_q)}\in[0,1]$$
where $\mathrm{diam}(R_q)$ is used as the normalising factor and represents the diameter of the shared support of $P_m$ and $P_s$ for a question $q\in Q^\mathrm{ord}$, i.e., the greatest possible distance between any two responses $r, r' \in R_q$ 
\begin{align}
    \nonumber \mathrm{diam}(R_q)  &= \sup_{r, r' \in R_q} |r - r'| = \max_{r, r' \in R_q} |r - r'| \\
    \nonumber &= \max(R_q) - \min(R_q).
\end{align}

\paragraph{Total variation distance} measures the largest absolute difference between the probabilities that the two distributions assign to the same event. It is applicable to all discrete distributions, whether ordinal or categorical and normalised and symmetric by default.

For each $q \in Q^\mathrm{nom}$, the total variation between the true and model response distributions ($P_m$ and $P_s$ respectively) is given by:
$$
d_{\mathrm{TV}}(P_m, P_s) \;:=\; \tfrac{1}{2} \sum_{r \in R_q} \big| P_m(r) - P_s(r) \big| \;\in[0,1].
$$
This can be understood as a special case of the Wasserstein distance, corresponding to optimal transport where the cost is zero when two categories coincide and one otherwise. In this sense, total variation is the natural analogue of Wasserstein for unordered response sets as it quantifies the minimum fraction of probability mass that must be reassigned across categories for the two distributions to align.

\subsection{Constructing the Correlation Matrices}\label{app:correlation}

In this analysis, we ask: \textit{do simulated responses reproduce the way in which items co-vary across demographic subgroups, as observed in the WVS?} To answer this, we construct and compare correlation matrices derived from the WVS and from each model configuration (OpinionGPT and persona prompting) and do so at two levels of granularity: \begin{enumerate*}[label=(\roman*)]
    \item the \textit{\textbf{question–question}} level, which considers correlations between individual survey questions; and 
    \item the \textit{\textbf{topic–topic}} level, where question are collapsed into the thematic value domains defined in the WVS. 
\end{enumerate*} The first tests whether models capture fine-grained associations between specific attitudes, while the second evaluates whether they reproduce the broader inter-domain relationships that underpin cultural values theories.

We first aggregate responses by computing the (normalised) \textit{\textbf{question mean responses}}\footnote{This calculation applies only to ordinal questions $Q^{ord}\subset Q$. Questions with nominal scales constitute less than 10\% of the survey, so the correlation structure analysis still covers the vast majority of the questionnaire.} in the matrix $\mathrm{A}^{\mathrm{WVS}}\in \mathbb{R}^{|Q^\mathrm{ord}|\times|S|}$ for the empirical WVS responses, with entries:
$$
A^{WVS}_{q,s} \;=\;\sum_{r\in R_q} \tilde{r}\, P_s(r\mid q) 
$$
the mean numerical response to item $q \in Q^\mathrm{ord}$ from subgroup $s \in S$ under the response distribution $P_s$, where $\tilde{r}:=\frac{r-\min(R_q)}{\mathrm{diam}(R_q)}$ represents the minmax normalised response value. This is done analogously with the corresponding model distributions $P_m$.

We then compute the pairwise Pearson correlation coefficients between all questions to produce the \textbf{\textit{question-question correlation matrix}} $\mathrm{C} \in \mathbb{R}^{|Q^\mathrm{ord}| \times |Q^\mathrm{ord}|}$ with elements
$$
C_{ij} = \mathrm{corr}\!\big(\mathrm{A}_{i,\cdot}, \, \mathrm{A}_{j,\cdot}\big)
$$
where $\mathrm{A}_{q,\cdot}$ denotes the vector of mean responses for question $q\in Q^\mathrm{ord}$ across all subgroups. To construct the \textbf{\textit{topic-topic correlation matrix}}, the same procedure applies with the extra step of first aggregating questions to the thematic domains before constructing correlations between topic-level response vectors.

\subsection{Bootstrapped Confidence Intervals}\label{app:correlation-bootstrap}
To estimate sampling variability for the metrics defined in \autoref{sec:correlation-eval}, we calculate bootstrapped confidence intervals. Given a model $m\in\{\text{OpinionGPT}, \text{Persona}\}$, which was used to simulate $500$ responses to each $q \in Q$ (denoted by $X^m$), we construct a 95\% confidence interval as follows:
\vspace{2mm}

\noindent For each iteration $b=1,...,B=1000$
\begin{enumerate}[leftmargin=*, noitemsep]
    \item Randomly resample $500$ responses ($\forall q\in Q$) with replacement from the generated responses. Let the bootstrap sample of responses be denoted by $X^{m, (b)}$
    \item Construct the bootstrap response distributions, $P_m^{(b)}$, from $X^{m, (b)}$
    \item Compute the correlation matrix from $P_m^{(b)}$ as in \autoref{sec:correlation-eval}
    $$\mathrm{C}^{m,(b)}\in \mathbb{R}^{|Q^{\mathrm{ord}}| \times |Q^{\mathrm{ord}}|}$$ 
    \item Compare $\mathrm{C}^{m,(b)}$ with the true WVS correlation matrix, $\mathrm{C}^{\mathrm{WVS}}$ according to the process outlined in \autoref{sec:correlation-eval} to yield the same two similarity metrics
    \begin{align}
    \nonumber c^{(b)} &= \mathrm{corr}\big(\mathrm{u}^{(b)}, \mathrm{u}^{\mathrm{WVS}}\big) \\
    \nonumber e^{(b)} &= \mathrm{RMSE}\big(\mathrm{u}^{(b)}, \mathrm{u}^{\mathrm{WVS}}\big)
    \end{align}
    where $\mathrm{u}$ once again denotes the vector obtained by stacking the upper triangle of a correlation matrix $\mathrm{C}$
\end{enumerate}
Finally, the confidence intervals are constructed from the 2.5th and 97.5th percentiles of the bootstrapped estimates $\{c^{(b)}\}_{b=1}^B$ and $\{e^{(b)}\}_{b=1}^B$, yielding the intervals
$[c_{0.025}, c_{0.975}]$ and $[e_{0.025}, e_{0.975}]$.

\subsection{Upper and Lower Bound Estimation for Correlation Structure Metrics}\label{app:correlation-lb-ub}

To contextualise the evaluation of correlation-structure similarity, we use two complementary procedures: \begin{enumerate*}[label=(\arabic*)]
    \item a \textit{\textbf{permutation-based null baseline}}, corresponding to the absence of cross-item correlation structure; and
    \item a \textbf{\textit{split-half resampling}} analysis, corresponding to an empirical "best case" where the compared correlation matrices are constructed from subsets of the same distribution
\end{enumerate*}. Under their respective scenarios, these provide pessimistic and optimistic estimates on the proposed metrics beyond the trivial bounds of the range of the RMSE and $\rho$ functions ([0, 1] and [-1, 1], respectively).

\paragraph{Permutation-based null baseline.}

While the unsteered \texttt{Phi-3} model serves as a baseline for marginal distribution comparisons, its subgroup-agnostic nature precludes its use for evaluating inter-subgroup correlations. We therefore construct a separate baseline via per-column (i.e., per question or topic) permutation of subgroup means, preserving the empirical distribution of each column's subgroup means while destroying any consistent subgroup ordering shared across questions. 

This yields a null distribution for correlation-structure similarity under the hypothesis of no shared inter-question dependence, against which model performance can be interpreted. Model performance can then be interpreted relative to this null: correlation structures that only marginally exceed permutation-based similarity indicate limited recovery of empirical inter-subgroup dependence.
\vspace{2mm}

\noindent For each iteration $b=1,...,B=1000$
\begin{enumerate}[leftmargin=*, noitemsep]
    \vspace{-2mm}
    \item For each column, $j$, of the empirical subgroup mean matrix $\mathrm{A}^{\mathrm{WVS}}\in \mathbb{R}^{|S|\times n}$, independently sample a random permutation of the column values to yield a permuted mean matrix $$\mathrm{A}^{(b)}\in\mathbb{R}^{|S|\times n}$$
    \item Use $\mathrm{A}^{(b)}$ to compute the corresponding correlation matrix $$\mathrm{C}^{(b)}_{\mathrm{null}}\in \mathbb{R}^{|Q^{\mathrm{ord}}| \times |Q^{\mathrm{ord}}|}$$
    \item Calculate the same similarity metrics between correlation matrices, $\mathrm{C}^{(b)}_{\mathrm{null}}$ and $\mathrm{C}^{\mathrm{WVS}}$
    \begin{align}
    \nonumber c^{(b)}_{\text{null}} &= \mathrm{corr}\big(\mathrm{u}^{(b)}_{\text{null}}, \mathrm{u}^{\mathrm{WVS}}\big) \\
    \nonumber e^{(b)}_{\text{null}} &= \mathrm{RMSE}\big(\mathrm{u}^{(b)}_{\text{null}}, \mathrm{u}^{\mathrm{WVS}}\big)
    \end{align}
\end{enumerate}
We take the mean values, $\bar{c}=\frac{1}{B}\sum^B_{b=1} c^{(b)}$ and $\bar{e}=\frac{1}{B}\sum^B_{b=1} e^{(b)}$, to define a floor on the correlation and a ceiling on the RMSE, respectively.

The above procedure is outlined for the question-question correlation structures. The same procedure applies for topic-topic level analysis with the additional step of first aggregating responses accordingly.

\paragraph{Split-half resampling analysis.} Having established a lower bound via permutation of subgroup means, we next estimate an upper bound on correlation-structure similarity by quantifying the intrinsic stability of the empirical WVS correlation matrix, $C^{\mathrm{WVS}}$, under respondent-level resampling. If $C^{\mathrm{WVS}}$ itself is noisy, then no model can reliably exceed that similarity level. To assess this, we apply the procedure outlined in \autoref{sec:correlation-eval} as part of a split-half correlation analysis. This then sets a ceiling on the correlation between a true and simulated correlation matrix and floor on the RMSE.
\vspace{2mm}

\noindent For each iteration $b=1,...,B=1000$
\begin{enumerate}[leftmargin=*, noitemsep]
    \vspace{-2mm}
    \item Within each of the ten subgroups, randomly split respondents into two halves
    \item For each half $h\in \{1, 2\} $, compute the subgroup means over each (ordinal-scaled) question 
    $$\mathrm{A}^{(b, h)}\in \mathbb{R}^{|S| \times |Q^{\mathrm{ord}}|}$$
    \item Use the means to construct each correlation matrix
    $$C^{(b,h)}\in \mathbb{R}^{|Q^{\mathrm{ord}}| \times |Q^{\mathrm{ord}}|}$$
    \item Compute the two metrics, RMSE and Pearson correlation, between the correlation matrices, i.e., between each half of the data
    \begin{align}
        \nonumber c^{(b)}_{1, 2} &= \mathrm{corr} \big(\mathrm{u}^{(b, 1)}, \, \mathrm{u}^{(b, 2)}\big)\\
        \nonumber e^{(b)}_{1,2} &= \mathrm{RMSE} \big(\mathrm{u}^{(b, 1)}, \, \mathrm{u}^{(b, 2)}\big)
    \end{align}
\end{enumerate}

Again, the resulting means for each metric are used to define the upper bound on model performance. Topic-topic analyses are constructed analogously as before.

\section{Supporting Results and Analyses}

\begin{figure*}[ht]
    \centering
    \includegraphics[width=0.48\textwidth]{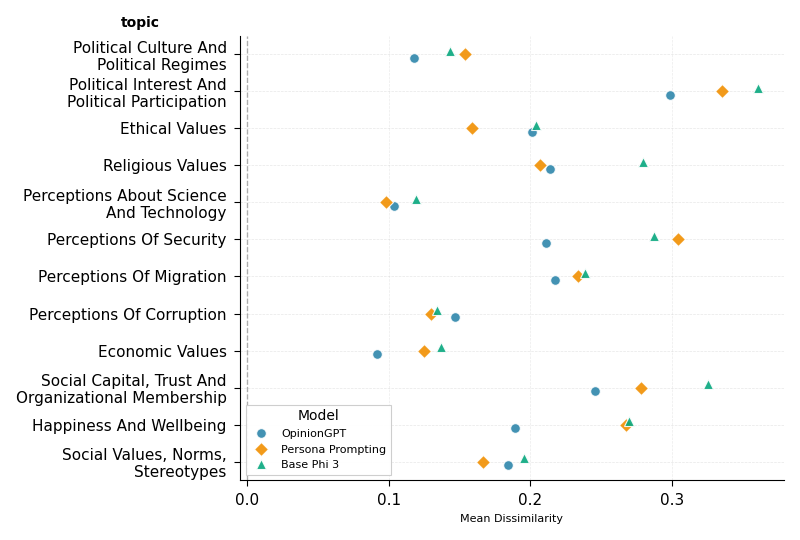}
    \includegraphics[width=0.48\textwidth]{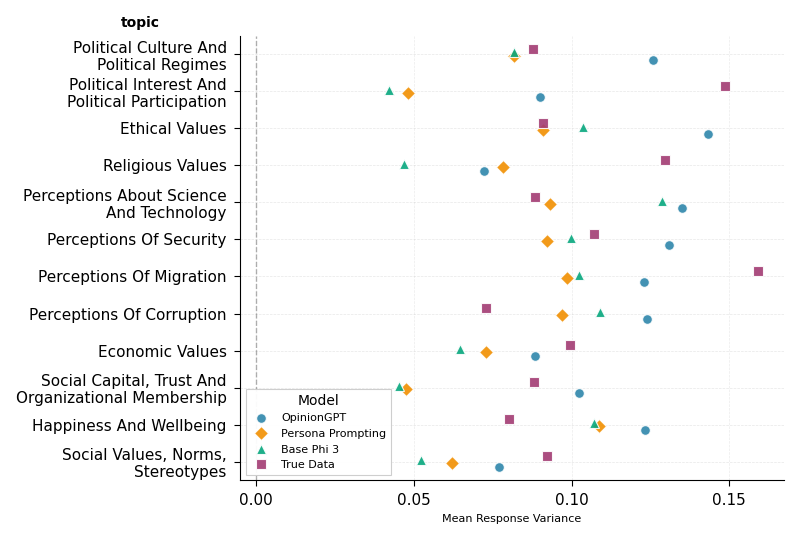}
    \caption{Evaluation metrics by question topic. Left: mean dissimilarity (lower = better). Right: mean response variance (closer to true data = better).}
    \label{fig:category-eval}
\end{figure*}

\subsection{Invalid Response Rates}\label{app:invalid}
Table \ref{tab:invalid_responses} shows the proportion of invalid responses produced by OpinionGPT and persona promoting for each demographic subgroup compared with the corresponding true non-response rates from WVS Wave 7. The true data contains comparatively fewer non-responses than either modeling approach, with OpinionGPT producing notably more invalid responses than persona prompting. Additionally, OpinionGPT displays much greater differences in invalid response rate across subgroups 

\begin{table}[]
    \centering
    \begin{tabular}{llll}
        \toprule
        Subgroup &
        \makecell[l]{True\\Responses} &
        \makecell[l]{Opinion\\GPT} &
        \makecell[l]{Persona\\Prompting}\\
        \midrule
        Liberal & 0.014 & 0.144 & \textbf{0.051} \\
        Conservative & 0.016 & 0.070 & \textbf{0.059} \\
        \midrule
        German & 0.024 & 0.070 & \textbf{0.054} \\
        American & 0.001 & 0.096 & \textbf{0.059} \\
        Latin America & 0.021 & 0.107 & \textbf{0.055} \\
        Middle East & 0.028 & 0.113 & \textbf{0.055} \\
        \midrule
        Men & 0.019 & \textbf{0.055} & 0.058 \\
        Women & 0.027 & 0.087 & \textbf{0.056} \\
        \midrule
        People Over 30 & 0.021 & 0.065 & \textbf{0.062} \\
        Old People & 0.025 & 0.091 & \textbf{0.061} \\
        \bottomrule
    \end{tabular}   
    \caption{Share of invalid responses per subgroup for the true responses, OpinionGPT and Persona Prompting.}
    \label{tab:invalid_responses}
\end{table}

Both OpinionGPT and persona prompted models, produced particularly high rates of invalid responses for questions with numeric response options, for which all OpinionGPT modules and persona-prompted models had >10\% invalid responses. This was a consistent trend for both model configurations across questions with similar numeric response sets. High rates of invalid responses were also observed on some questions beyond this, although without a clear structural pattern to the questions.

\subsection{Modal Collapse}\label{app:collapse}

\begin{table}[h]
\small
\centering
\begin{tabular}{lrr}
\toprule
 & OpinionGPT & Persona Prompting \\
\midrule
Liberal & \textbf{6} & 17 \\
Conservative & \textbf{6} & 10 \\
German & \textbf{7} & 13 \\
American & \textbf{0} & 11 \\
Latin America & \textbf{0} & 6 \\
Middle East & \textbf{0} & 10 \\
Men & \textbf{0} & 7 \\
Women & \textbf{0} & 9 \\
People Over 30 & \textbf{2} & 8 \\
Old People & \textbf{4} & 7 \\
\bottomrule
\end{tabular}

\caption{No. of questions with only a single response by model}
\label{tab:degenerate-counts}
\end{table}

A common consequence of insufficient response diversity is the collapse of the output distribution towards the mode. \autoref{tab:degenerate-counts} shows the incidence (by number of question) of this \textit{\textbf{modal collapse}} for each subgroup. It can be seen that the OpinionGPT modules result in much less modal collapse than persona prompted models; for the unsteered baseline there were $10$ questions with modal collapse.

\subsection{Effect of Omitting Categorical Questions}\label{app:categorical-omission}

In order to construct the correlation matrices (as outlined in Appendix~\autoref{app:correlation}), we first calculate the mean responses. As categorical or nominal response scales lack the notion of a mean response, we leave them out of this analysis. This omits $18$ questions of our data subset, or less than $10\%$ of the $193$ questions, thus leaving the majority of questions covered by the comparison of correlation structures.

\begin{figure*}[tb!]
    \centering
    \includegraphics[width=0.48\textwidth]{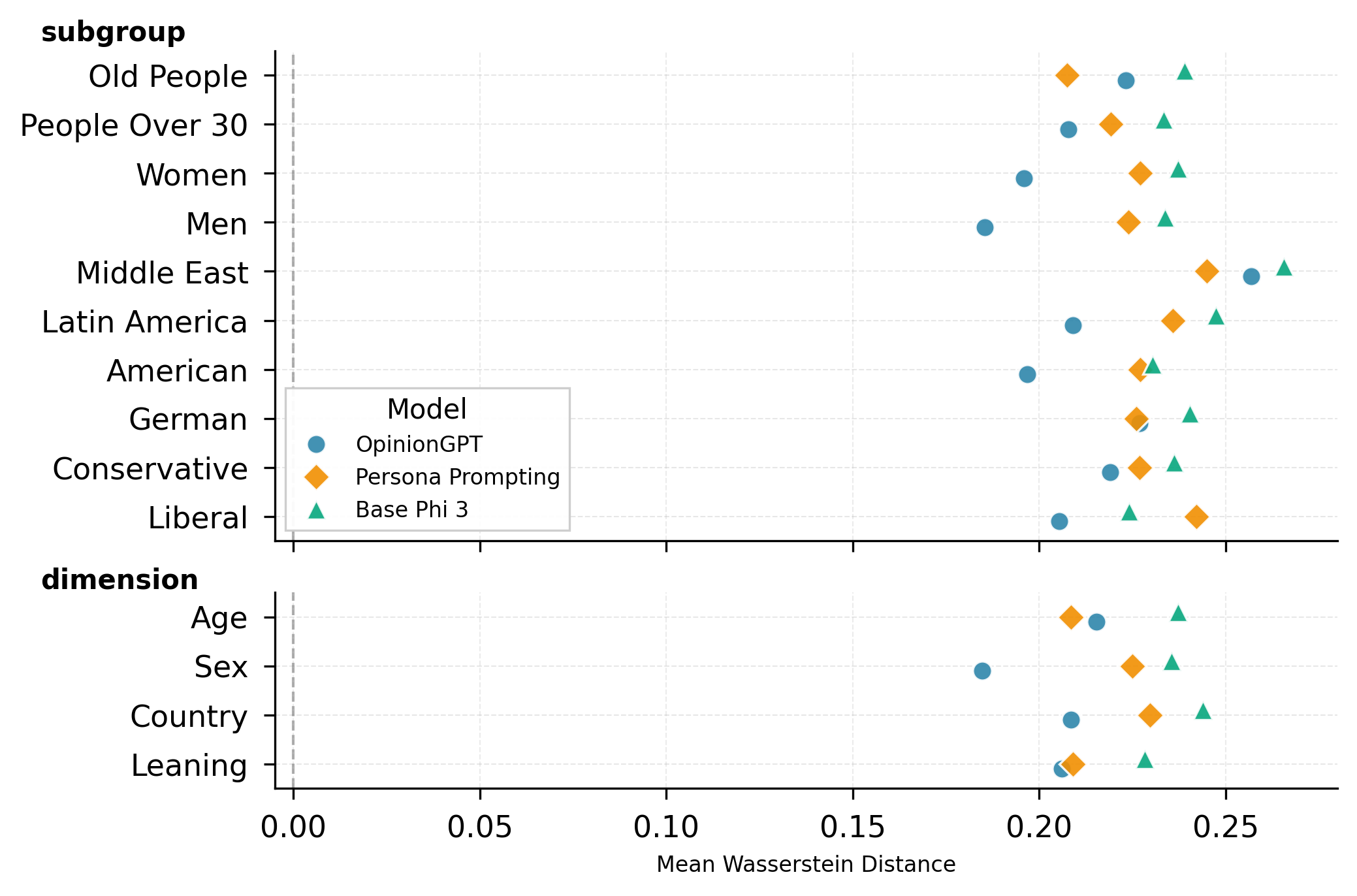}
    \includegraphics[width=0.48\textwidth]{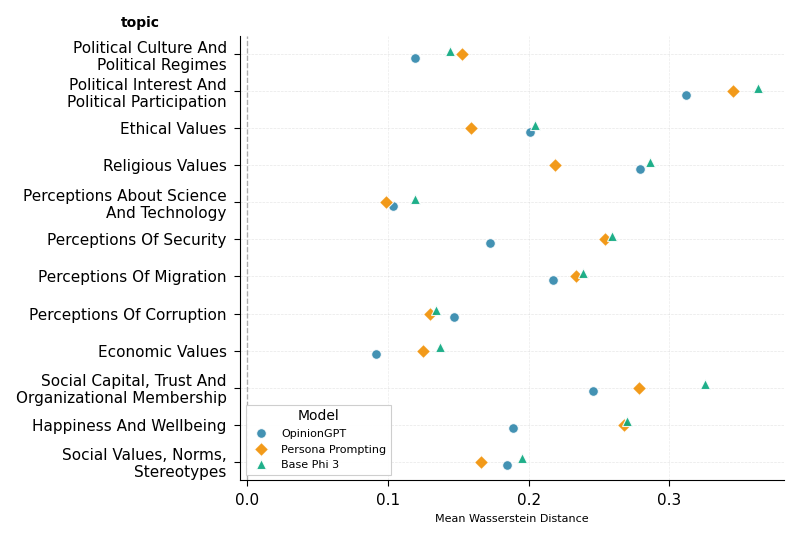}
    \caption{Mean dissimilarity without categorical questions (lower = better). Left: by demographic subgroup and dimension. Right: by question topic}
    \label{fig:wasserstein-eval}
\end{figure*}

To see the effect this omission would have on marginal dissimilarity scores we can calculate the Wasserstein distances for questions with Likert-type/numerical response scales, shown in \autoref{fig:wasserstein-eval}. Compared to the complete score by subgroup in \autoref{sec:marginal-eval}, omitting categorical questions only changes dissimilarity scores minimally, $\pm0.01$ or around $1~$--$~4\%$. For the question topics only four\footnote{\textit{Perceptions of Security}, \textit{Religious Values}, \textit{Political Interest and Political Participation} and \textit{Political Culture and Political Regimes}} have questions with categorical scales with moderate differences for \textit{Perceptions of Security} and \textit{Religious Values} and minimal or no differences for the remaining topics.

\label{sec:experiment-dists}

\section{Ablations}
\subsection{Temperature Effects}\label{app:temp}
Sampling temperature is a key decoding parameter that controls the entropy of the token distribution and, consequently, the variability of responses generated by large language models

Sampling temperature is a key decoding parameter that controls the entropy of the model's token-level output distribution and, consequently, directly influences the variability of responses generated by large language models \citep{holtzman_curious_2020}. Lower values make the model more deterministic, whereas higher values increase response diversity but may produce erratic or unfaithful outputs. There is no clear, natural value for the temperature that guarantees optimal results. We therefore test the model's marginal dissimilarity and response variance for different temperature settings. The results are in table \ref{tab:temperature}.

\begin{table}[]
    \centering
    \begin{tabular}{c|c|c}
        \toprule
        Temp. & OpinionGPT & Persona Prompting \\
        \midrule
        0.6   & 0.223 \textit{(0.104)} & 0.235 \textit{(0.068)}     \\
        0.7   & 0.214 \textit{(0.106)} & 0.230 \textit{(0.070)}     \\
        0.8   & 0.205 \textit{(0.108)} & 0.224 \textit{(0.073)}     \\
        0.9   & 0.198 \textit{(0.109)} & 0.219 \textit{(0.075)}     \\
        1.0   & 0.192 \textit{(0.112)} & 0.213 \textit{(0.078)}  \\
         \bottomrule
    \end{tabular}
    \caption{Average marginal dissimilarity and variance (in parentheses) for persona prompting and OpinionGPT across different temperature settings, averaged across all four sociodemographic dimensions.}
    \label{tab:temperature}
\end{table}

The results highlight a trade-off between variance, alignment, modal collapse, and validity. As temperature increased, the variance of both OpinionGPT and persona-prompted responses increased. However, the persona-prompted Phi-3 remained consistently under-dispersed even at high temperature values. Alignment improved at higher temperatures, indicating that moderate stochasticity better captures the population-level patterns. At low temperatures, both models suffered from modal collapse, with several survey items converging onto degenerate distributions, a problem that was substantially alleviated at higher settings (see also Appendix \ref{app:collapse}. However, this improvement came alongside a marked increase in invalid responses, particularly pronounced in OpinionGPT relative to persona prompting. The choice of temperature therefore involves a fundamental methodological compromise. 

For our evaluation, we choose a temperature of 0.9 in light of improved alignment and low model collapse, while still controlling the invalid response rate to an extent. However, although the exact balance between modal collapse and invalid response rates varies with the choice of temperature, the broader conclusions of this study, concerning the relative dispersion, alignment, and representativeness of the models, remain robust across settings.
\subsection{Response Order Effects}\label{app:response_order}
To test for response order effects, we randomly flip the response scale for 50\% of samples.  We then prompt the model with the questions and these original and flipped response scales, normalise the response options to a common [0,1] scale and compare the resulting output distributions by computing per-item means. Table \ref{tab:response_order} shows these means.

\begin{table}[!h]
    \centering
    \begin{tabular}{l|ccc|ccc}
    \toprule
    Subgroup & \multicolumn{3}{l}{Persona Prompting} & \multicolumn{3}{l}{OpinionGPT} \\
         & orig          & flip    & diff      & orig       & flip  & diff   \\
        \midrule
        Liberal & 0.48 & 0.26 & \textit{0.22} & 0.46 & 0.32 & \textit{0.14} \\
        Conservative & 0.44 & 0.27 & \textit{0.17} & 0.45 & 0.32 & \textit{0.13} \\
        \midrule
        Germany & 0.48 & 0.27 & \textit{0.21} & 0.50 & 0.32 & \textit{0.17} \\
        America & 0.47 & 0.26 & \textit{0.21} & 0.49 & 0.32 & \textit{0.17} \\
        Latin America & 0.47 & 0.26 & \textit{0.21} & 0.51 & 0.34 & \textit{0.16} \\
        Middle East & 0.46 & 0.25 & \textit{0.20} & 0.52 & 0.25 & \textit{0.28} \\
        \midrule
        Men & 0.47 & 0.26 & \textit{0.21} & 0.55 & 0.30 & \textit{0.25} \\
        Women & 0.48 & 0.26 & \textit{0.22} & 0.54 & 0.26 & \textit{0.29} \\
        \midrule
        P. Over 30 & 0.47 & 0.27 & \textit{0.20} & 0.55 & 0.27 & \textit{0.28} \\
        Old People & 0.49 & 0.27 & \textit{0.23} & 0.56 & 0.26 & \textit{0.30} \\
        \bottomrule
    \end{tabular}
    \caption{[0,1]-normalised mean responses for persona prompting and OpinionGPT, with the original (orig) and flipped (flip) order of the answer options and their difference (diff). }
    \label{tab:response_order}
\end{table}

The results revealed substantial response order effects across all models, with an average difference of approximately 0.20 between the normalised mean responses under the original and flipped orderings (after remapping to the original scale). In every subgroup, the original ordering produced systematically higher values, consistent with a tendency to favour later options in the list, akin to a recency bias. Although recency effects are recorded more frequently than primacy effects in human surveys, the effects are generally modest \citep{krosnick_evaluation_1987, holbrook_response_2007}. We find some differences between persona prompting and OpinionGPT with persona prompting having more consistent response order effects, but they are not systematic, meaning that these response order effects are likely a result of the base model and not of the finetuned adapters or the persona prompt. One reason for these strong order effects may be the small size of our base model (\texttt{Phi-3-mini-Instruct} with 3.5B parameters).

To reduce this response-order biases in our findings, we keep the flipped response order for 50\% of the sampled responses to each question and retain the original order for the other 50\%. For the evaluation, we reassign the flipped responses to the original scale.
\end{document}